\documentclass{article} 
\usepackage{preprint,times}

\usepackage{amsmath,amsfonts,bm}

\def\eqref#1{equation~\ref{#1}}

\def\1{\bm{1}}

\DeclareMathAlphabet{\mathsfit}{\encodingdefault}{\sfdefault}{m}{sl}
\SetMathAlphabet{\mathsfit}{bold}{\encodingdefault}{\sfdefault}{bx}{n}

\usepackage{titletoc}
\usepackage{caption}
\usepackage{hyperref}
\hypersetup{
    hidelinks,
    pdftitle={Understanding Mamba-3 Through Oscillatory Dynamics},
    pdfsubject={Oscillatory dynamics in recurrent sequence models},
    pdfkeywords={Mamba-3, state-space models, oscillation, long-context modeling}
}
\usepackage{url}
\usepackage{tabularx}
\usepackage{booktabs}
\usepackage{wrapfig}
\usepackage{graphicx}
\usepackage{placeins}
\usepackage{float}
\usepackage{needspace}
\usepackage{multirow}
\usepackage{enumitem}
\usepackage{makecell}

\usepackage[most]{tcolorbox}
\definecolor{findingteal}{HTML}{147D92}

\newtcolorbox{findingbox}{
    enhanced,
    colback=findingteal!5!white,
    colframe=findingteal,
    frame hidden,
    borderline west={2pt}{0pt}{findingteal},
    arc=1mm,
    boxrule=0pt,
    left=9pt,
    right=8pt,
    top=6pt,
    bottom=6pt,
    before skip=8pt,
    after skip=8pt
}

\makeatletter
\newcommand{\finishwrap}{%
    \par
    \ifnum\value{WF@wrappedlines}>1\relax
        \vspace{\dimexpr
            \value{WF@wrappedlines}\baselineskip
            -\baselineskip\relax}%
    \fi
    \WFclear
}
\makeatother

\title{Understanding Mamba-3 Through Oscillatory Dynamics}

\author{%
  \textbf{Shutian Liang}, \textbf{Yue Song}\thanks{Denotes corresponding author.} \\
  College of AI, Tsinghua University
}
\hypersetup{pdfauthor={Shutian Liang, Yue Song}}

\date{}

\begin{document}

\maketitle

\begin{abstract}

Mamba-3 introduces input-dependent rotations into recurrent state updates, but their contribution beyond state retention remains unclear. We interpret these rotations as \textbf{underdamped second-order dynamics}, providing a framework for investigating the role of oscillation in sequence modeling. Controlled comparisons show that underdamped dynamics outperform non-oscillatory damping regimes on state-tracking tasks, while language-model pretraining demonstrates benefits for long-context prediction and length generalization. Interventions in pretrained models further establish the importance of learned phase, and state analysis suggests that rotation regulates how retained information combines with incoming updates. Taken together, these findings identify underdamped dynamics as a useful mechanism for recurrent computation and suggest that oscillation supports long-context processing by regulating how information accumulates. The project and code are available at {\urlstyle{same}\url{https://shutian-liang.github.io/oscillationmamba/}}.

\end{abstract}

\section{Introduction}

State-space models (SSMs) process sequences through recurrent updates to a hidden state~\citep{gu2021efficiently}. Their dynamical structure enables linear-time sequence processing and constant-memory autoregressive inference, while structured algorithms support parallel training. These properties make SSMs an efficient alternative to attention-based models~\citep{vaswani2017attention}. In selective SSMs such as Mamba-1~\citep{gu2023mamba} and Mamba-2~\citep{dao2024transformers}, input-dependent updates determine what information enters and leaves the state, while multiplicative decay controls how much of the existing state is retained.

\begin{figure}[t]
    \centering
    \includegraphics[width=0.9\linewidth]{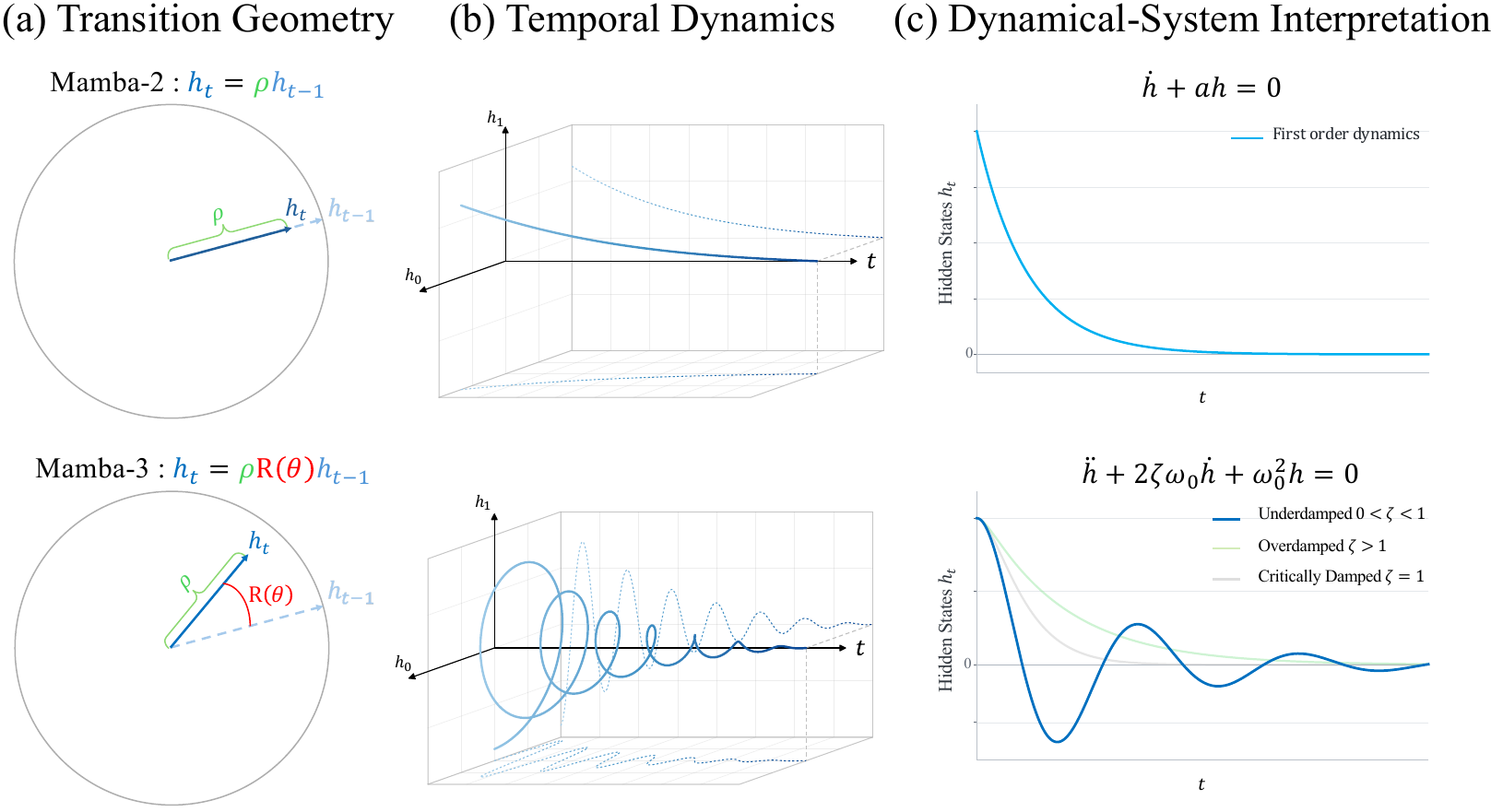}
    \caption{
    \textbf{A geometric perspective on Mamba-3's recurrent dynamics.}
    (a) Mamba-2 scales the retained state, while Mamba-3 additionally rotates it.
    (b) With fixed coefficients and no input, decay--rotation produces a spiraling trajectory.
    (c) This geometry motivates our interpretation of Mamba-3 as an underdamped second-order system with oscillatory dynamics.
    }
    \label{fig:teaser}
    \vspace{-16pt}
\end{figure}

Mamba-3~\citep{lahoti2026mamba} adds input-dependent rotation to this state evolution through complex-valued transitions. Unlike conventional rotary positional embeddings in Transformers~\citep{su2024roformer}, whose angles follow a prescribed positional schedule, Mamba-3 learns phase increments from the input. As illustrated in Fig.~\ref{fig:teaser}, these rotations change the direction of the retained state, turning pure decay into spiraling evolution under fixed coefficients. This geometric distinction raises a natural computational question: \emph{what do learned rotations contribute beyond state retention, particularly over long contexts?}

We investigate this question by interpreting Mamba-3's decay--rotation transition as an \textbf{underdamped second-order system}. In this representation, decay sets damping envelopes, while rotation produces oscillatory phase evolution. The connection provides a unified framework for comparing underdamped, critically damped, and overdamped dynamics. Our analysis characterizes how these regimes differ in mode persistence and state geometry, and controlled state-tracking experiments show a clear advantage of underdamped dynamics. Since all three variants retain coupled second-order states, the comparisons highlight the computational contribution of the oscillatory regime.

Our language-modeling experiments examine whether this contribution extends to the use of contextual information. In pretrained Mamba-3 models, removing phase rotation affects context-dependent prediction on LAMBADA~\citep{paperno2016lambada} much more strongly than factual relation prediction~\citep{hernandez2024linearity}. We further train oscillatory and non-oscillatory models from scratch to assess how rotation affects learned long-context capabilities. Experimental results show that oscillatory models achieve lower PG-19 perplexity~\citep{rae2019compressive} beyond the training context length at both evaluated scales, with downstream gains on LongBench-E~\citep{bai2024longbench} at the larger scale.

To understand how oscillation contributes to these gains, we examine how learned phase shapes recurrent computation through interventions in pretrained Mamba-3 models. Removing, reversing, or shuffling phase increments substantially degrades long-context prediction. Phase removal also increases alignment between retained history and incoming updates, together with larger state norms, suggesting that rotation moderates constructive accumulation. Interventions on writing and readout further reveal the importance of compatible phase coordinates for accessing retained information.

Our main findings are:
\begin{itemize}
\item \textbf{Underdamped dynamics provide a useful interpretation of Mamba-3.}
The second-order representation connects its learned rotations to oscillation and enables damping-regime comparisons that demonstrate the benefit of underdamping for state tracking.

\item \textbf{Oscillation benefits long-context language modeling.}
Controlled pretraining shows improved length generalization at both evaluated scales (187M and 444M), with downstream long-context gains emerging at the larger scale.

\item \textbf{Learned phase regulates recurrent state accumulation.}
Phase interventions establish its importance for long-context prediction, while state analysis suggests that rotation moderates constructive accumulation between retained history and incoming updates.
\end{itemize}

Together, these findings suggest that effective recurrent memory in SSMs depends on both how much information is retained and how that information is transformed as new inputs arrive. More broadly, this dynamical perspective motivates oscillatory dynamics as a design dimension for sequence architectures, with extensions to linear-attention and other recurrent models (Appendix~\ref{app:extension}). We further discuss state dynamics as a design principle in our blog post, \href{https://shutian-liang.github.io/backtodynamics/}{\emph{Back to Dynamics}}.


\section{Background}
\subsection{Selective State-Space Models}
\label{ssm}

State-space models process sequences by maintaining a recurrent hidden state. In continuous time, their dynamics take the form
\begin{equation}
    \dot h_t=A_th_t+B_tx_t,
    \qquad
    y_t=C_t^\top h_t,
\end{equation}
where $x(t),y(t)\in\mathbb{R}^{d}$ and $h(t)\in\mathbb{R}^{h}$. The transition matrix $A(t)\in\mathbb{R}^{h\times h}$ governs the evolution of the existing state, while $B(t),C(t)\in\mathbb{R}^{h\times d}$ control input injection and output readout. Discretizing over intervals of length $\Delta_t$ yields a recurrence of the form
\begin{equation}
    h_t=\bar A_t h_{t-1}+\bar B_t x_t,
    \qquad
    y_t=C_t^\top h_t.
\end{equation}
Selective SSMs make components of this recurrence input-dependent, allowing each token to influence how information is retained, written, and read. We focus on the transition $\bar A_t$, whose structure distinguishes the state dynamics of Mamba-1, Mamba-2, and Mamba-3.

\noindent\textbf{Mamba-1: selective diagonal decay.}
Mamba-1 makes $\Delta_t$, $B_t$, and $C_t$ input-dependent while retaining a learned, time-invariant diagonal transition matrix $A$. Its discrete transition is
\begin{equation}
    \bar A_t=\exp(\Delta_t A),
\end{equation}
With negative diagonal entries in $A$, each state coordinate decays at its own rate, and the input-dependent step size $\Delta_t$ modulates these rates across tokens.

\noindent \textbf{Mamba-2: shared decay within each head.} Mamba-2 restricts the continuous-time transition within each head to $A=aI$, where $a<0$. The discrete update therefore becomes
\begin{equation}
\bar A_t=\alpha_t I,
\qquad
\alpha_t=\exp(\Delta_t a),
\qquad
h_t=\alpha_t h_{t-1}+\Delta_t B_t x_t.
\end{equation}
The scalar $\alpha_t$ controls retention uniformly across state coordinates within a head. Consequently, homogeneous propagation changes the magnitude of the state while preserving its direction, while token-dependent writing and readout remain controlled by $B_t$ and $C_t$.

\noindent\textbf{Mamba-3: decay with input-dependent rotation.} Mamba-3 introduces a phase degree of freedom that allows the propagated state to change direction as well as decay. For each two-dimensional rotary state pair, the transition takes the form
\begin{equation}
    \bar A_t=\rho_t R(\theta_t),
    \qquad
    R(\theta_t)=
    \begin{bmatrix}
        \cos\theta_t & -\sin\theta_t \\
        \sin\theta_t &  \cos\theta_t
    \end{bmatrix},
    \label{eq:R}
\end{equation}
where $\rho_t$ controls radial retention and $\theta_t$ is an input-dependent rotation angle, implemented through data-dependent rotary embeddings. Mamba-3's single-input, single-output (SISO) and multi-input, multi-output (MIMO) variants share this decay--rotation structure but differ in their input and readout parameterizations. Our analysis mainly focuses on the dynamical role of the learned phase.

\subsection{Damped Second-Order Dynamics}
A damped second-order system evolves according to
\begin{equation}
    \ddot x(t)
    +2\zeta\omega_0\dot x(t)
    +\omega_0^2 x(t)
    =u(t),
\end{equation}
where $\omega_0>0$ is the natural frequency, $\zeta>0$ is the damping ratio, and $u(t)$ is an external forcing term. Introducing the velocity $v(t)=\dot x(t)$ gives the first-order state-space representation
\begin{equation}
    \frac{d}{dt}
    \begin{bmatrix}
        x(t)\\
        v(t)
    \end{bmatrix}
    =
    \underbrace{
    \begin{bmatrix}
        0 & 1\\
        -\omega_0^2 & -2\zeta\omega_0
    \end{bmatrix}}_{A_{\mathrm{SO}}}
    \begin{bmatrix}
        x(t)\\
        v(t)
    \end{bmatrix}
    +
    \begin{bmatrix}
        0\\
        1
    \end{bmatrix}u(t).
\end{equation}
The eigenvalues of $A_{\mathrm{SO}}$ determine the system's intrinsic response:
\begin{equation}
    \lambda_{\pm}
    =
    -\zeta\omega_0
    \pm \omega_0\sqrt{\zeta^2-1}.
\end{equation}
Table~\ref{tab:damping_regimes} summarizes three regimes: underdamped systems have complex-conjugate eigenvalues and exhibit decaying oscillations, whereas critically damped and overdamped systems have real eigenvalues and non-oscillatory intrinsic dynamics. This distinction allows us to examine the role of oscillation within a common second-order framework.

\begin{table}[htbp]
    \centering
    \caption{Damping regimes of a second-order dynamical system.}
    \label{tab:damping_regimes}
    \begin{tabular}{lcc}
        \toprule
        Regime & Eigenvalues & Dynamics \\
        \midrule
        Underdamped ($0<\zeta<1$)
        & Complex conjugate & Oscillatory \\
        Critically damped ($\zeta=1$)
        & Repeated real & Non-oscillatory \\
        Overdamped ($\zeta>1$)
        & Distinct real & Non-oscillatory \\
        \bottomrule
    \end{tabular}
\end{table}



\section{An Oscillatory Interpretation of Mamba-3}
\label{sec:second_order_rope}

We now derive a second-order representation of Mamba-3's decay--rotation transition and identify its underlying damping regime. The analysis considers homogeneous state propagation, separating the intrinsic transition dynamics from input injection.

\noindent\textbf{From discrete rotation to continuous dynamics.}
Consider a rotary state pair $z_t=(H_t,P_t)^\top$ evolving as
\begin{equation}
    z_{t+1}
    =
    \rho_t R(\theta_t) z_t,
    \qquad
    0<\rho_t<1,
\end{equation}
where $R(\theta_t)$ is the two-dimensional rotation matrix defined in Eq.~\ref{eq:R}. For an interval of length $\Delta_t>0$, define the decay rate and angular frequency as
\begin{equation}
a_t=-\frac{\log\rho_t}{\Delta_t},
\qquad
\nu_t=\frac{\theta_t}{\Delta_t}.
\end{equation}
The corresponding continuous-time generator is
\begin{equation}
    A_t^{\mathrm{rot}}
    =
    -a_t I+\nu_t J,
 \qquad
    J=
    \begin{bmatrix}
        0 & -1 \\
        1 & 0
    \end{bmatrix},
\end{equation}
Since $J^2=-I$, its exponential satisfies
\begin{equation}
\label{eq:rotation_exponential}
e^{\phi J}=\cos(\phi)I+\sin(\phi)J=R(\phi),
\end{equation}
as derived in Appendix~\ref{app:rotation_exponential}. Because $I$ and $J$ commute,
\begin{equation}
    \begin{aligned}
        e^{A_t^{\mathrm{rot}}\Delta_t}
        &=
        e^{-a_t\Delta_t}
        e^{\nu_t\Delta_t J} \\
        &=
        e^{-a_t\Delta_t}
        \left[
            \cos(\nu_t\Delta_t)I
            +
            \sin(\nu_t\Delta_t)J
        \right] \\
        &=
        e^{-a_t\Delta_t}
        \begin{bmatrix}
            \cos\theta_t & -\sin\theta_t \\
            \sin\theta_t & \cos\theta_t
        \end{bmatrix}
        = \rho_t R(\theta_t).
    \end{aligned}
\end{equation}
Thus, each discrete decay--rotation update is the exact flow of $\dot z=A_t^{\mathrm{rot}}z$, with coefficients held fixed over that interval.

\noindent\textbf{An underdamped second-order representation.} Within one interval, suppress the index $t$ and write the continuous dynamics componentwise:
\begin{equation}
    \label{int:mamba3}
    \dot H=-aH-\nu P,
    \qquad
    \dot P=\nu H-aP.
\end{equation}
Differentiating the first equation and substituting the second yields
\begin{equation}
    \ddot H
    =
    -a\dot H-\nu(\nu H-aP)
    =
    -2a\dot H-(a^2+\nu^2)H,
\end{equation}
where we used $\nu P=-\dot H-aH$. Hence,
\begin{equation}
    \label{equ:secondorder}
    \ddot H+2a\dot H+(a^2+\nu^2)H=0.
\end{equation}
For $\nu\neq0$, the auxiliary coordinate is recovered as $P=-(\dot H+aH)/\nu$, so the two representations are equivalent within the interval.

Comparing with the canonical damped second-order equation gives
\begin{equation}
    \omega_0=\sqrt{a^2+\nu^2},
    \qquad
    \zeta=\frac{a}{\sqrt{a^2+\nu^2}},
    \qquad
    \omega_d=\omega_0\sqrt{1-\zeta^2}=|\nu|.
\end{equation}
Since $a>0$, any nonzero $\nu$ gives $0<\zeta<1$, placing the dynamics in the underdamped regime. The decay rate $a$ determines the damping envelope, while $|\nu|$ sets the oscillation frequency and the sign of $\nu$ specifies the rotation direction. Mamba-3's learned phase therefore controls the oscillatory component of state evolution. In Sec.~\ref{sec:reparameterization}, we compare this regime with critically damped and overdamped alternatives to assess the contribution of oscillation to recurrent computation.

\section{Oscillation and State Tracking}

Does the benefit of Mamba-3's rotary transition arise from oscillation or simply from coupling two state coordinates? We investigate this question by comparing oscillatory and non-oscillatory variants with the same second-order state structure. State-tracking tasks provide a controlled setting in which models must maintain and update an internal state as new tokens arrive.

\subsection{Controlled Comparisons Across Damping Regimes}
\label{sec:reparameterization}

To construct these variants, we generalize second-order dynamics in Eq.~\ref{equ:secondorder} through a parameter $\kappa_t$:
\begin{equation}
    A_t(\kappa_t)
    =
    \begin{bmatrix}
        0 & 1\\
        -(a_t^2-\kappa_t) & -2a_t
    \end{bmatrix},
    \label{eq:general_second_order}
\end{equation}
Within each interval, this generator corresponds to
\begin{equation}
    \ddot H
    +
    2a_t\dot H
    +
    (a_t^2-\kappa_t)H
    =
    U_t,
    \label{eq:general_second_order_scalar}
\end{equation}
where $U_t$ denotes an external forcing term. Its eigenvalues are:
\begin{equation}
    \label{eq:general_second_order_eigs}
    \lambda_{\pm}
    =-a_t\pm\sqrt{\kappa_t}.
\end{equation}
Thus, $\kappa_t$ controls the modal structure around the shared spectral center $-a_t$. Setting $\kappa_t=-\nu_t^2<0$ recovers the underdamped representation derived for Mamba-3 within each interval, while $\kappa_t=0$ and $0<\kappa_t<a_t^2$ yield critically damped and overdamped alternatives, respectively. 

All variants retain the coupled two-dimensional state structure. We use the same input-dependent step size and discretize each generator as
\begin{equation}
\bar A_t(\kappa_t)=\exp\!\left(\Delta_t A_t(\kappa_t)\right).
\end{equation}
The state dimension, input and readout, and input-update rule are kept unchanged across variants. This comparison tests the benefit of oscillatory modes beyond second-order state coupling alone.

\subsection{Mode Persistence and State Geometry}
\label{sec:eigenmode_memory}

We next examine how each damping regime affects the persistence and geometry of propagated states. We consider homogeneous dynamics with fixed coefficients, holding the natural frequency $\omega_0>0$ constant while varying $a>0$ and setting $\kappa=a^2-\omega_0^2$. The continuous-time generator and its propagation operator are
\begin{equation}
A(a)=
\begin{bmatrix}
    0 & 1\\
    -(a^2 - \kappa) & -2a
\end{bmatrix},
\qquad
\Phi_a(t)=e^{A(a)t},
\end{equation}

\noindent\textbf{Overdamped dynamics: dominance of a single mode.} For $a>\omega_0$, we have $\kappa>0$ and eigenvalues $\lambda_\pm=-a\pm\sqrt{\kappa}$. With a fixed step size $\Delta>0$, the discrete eigenvalues $\mu_\pm=e^{\lambda_\pm\Delta}$ satisfy
\begin{equation}
    \lim_{n\to\infty}
    |\frac{\mu_-}{\mu_+}|^n
    =
    \lim_{n\to\infty}e^{-2\sqrt{\kappa}\,n\Delta}
    =0.
\end{equation}
The fast mode therefore becomes negligible relative to the slow mode as the number of recurrent steps $n$ increases. For initial states with a nonzero slow-mode component, propagation becomes dominated by the slow eigendirection, indicating \emph{asymptotic mode collapse}.

\noindent\textbf{Critically damped dynamics: fastest asymptotic attenuation.} At $a=\omega_0$, the eigenvalues coincide at $-\omega_0$. Writing $A=-\omega_0 I+N$, where $N^2=0$ and $N\neq0$, gives
\begin{equation}
\Phi_{\omega_0}(t)=e^{-\omega_0t}(I+tN).
\end{equation}
The polynomial factor does not change the asymptotic exponential decay rate, which is $\omega_0$. By comparison, the dominant overdamped mode decays at rate
\begin{equation}
a-\sqrt{a^2-\omega_0^2}
=
\frac{\omega_0^2}{a+\sqrt{a^2-\omega_0^2}}
<\omega_0,
\end{equation}
while the underdamped decay rate is $a<\omega_0$. Critical damping therefore yields the \emph{fastest asymptotic exponential attenuation} at a fixed natural frequency. Its propagation also becomes directionally degenerate, since $e^{\omega_0t}\Phi_{\omega_0}(t)/t\to N$, a rank-one matrix.

\noindent\textbf{Underdamped dynamics: persistent two-dimensional geometry.}
For $0<a<\omega_0$, let $\nu=\sqrt{\omega_0^2-a^2}$. The eigenvalues are $\lambda_\pm=-a\pm i\nu$, giving
\begin{equation}
    |\mu_+|=|\mu_-|=e^{-a\Delta},
    \qquad
    \lim_{n\to\infty}
    |\frac{\mu_-}{\mu_+}|^n=1.
\end{equation}
Neither mode is preferentially attenuated. In the rotation coordinates, propagation takes the form
\begin{equation}
z_n=e^{-an\Delta}R(n\nu\Delta)z_0.
\end{equation}
Thus, lengths and angles are preserved up to a common radial decay, without progressive directional collapse. Since $a<\omega_0$, attenuation is also slower than under critical damping. Underdamped dynamics therefore combine \emph{preservation of relative state geometry} with \emph{slower asymptotic decay than critical damping}. Full proofs of mode persistence analysis are provided in Appendix~\ref{app:mode_persistence}.

\begin{wraptable}[11]{R}{0.5\linewidth}
    \vspace{-20pt} 
    \centering
    \caption{Scaled Accuracy (\%) of state-tracking.}
    \label{tab:state_tracking}
    \scriptsize
    \setlength{\tabcolsep}{2.2pt}
    \renewcommand{\arraystretch}{1.05}

    \resizebox{\linewidth}{!}
    {%
    \begin{tabular}{lccc}
        \toprule
        Model
        & Parity
        & Mod 3
        & \makecell{Arith. w/o\\Brackets} \\
        \midrule

        Mamba-2
        & 3.91
        & 0.04
        & 36.76
        \\

        Mamba-3 w/o rotation
        & 3.03
        & 0.78
        & 48.95 \\

        Mamba-3 w/ Static rotation
        & 4.05
        & 0.85
        & 25.49 \\


        \midrule

        Overdamped
        & 3.27
        & 0.58
        & 81.60 \\

        Critically Damped
        & 3.22
        & 0.32
        & 1.97 \\

        Underdamped (Mamba-3)
        & \textbf{100.0}
        & \textbf{95.26}
        & \textbf{95.68} \\
        \bottomrule
    \end{tabular}%
    }
\end{wraptable}



\subsection{State-Tracking Tasks}

\leavevmode\noindent\textbf{Experimental Setup.} We evaluate Parity, Mod 3, and Arithmetic without Brackets through two comparisons: varying the damping regime within the shared second-order framework, comparing Mamba-3 with fixed and absent rotations. These comparisons assess the importance of underdamped and input-dependent phase. We also include Mamba-2 as a baseline. 

\noindent\textbf{Results.} Table~\ref{tab:state_tracking} shows a consistent advantage of underdamped dynamics over critically damped and overdamped alternatives. The gap is particularly pronounced on Parity and Mod 3, whose cyclic structure naturally accommodates rotational updates. The advantage extends to Arithmetic without Brackets, where the underdamped variant achieves 95.68\% scaled accuracy, compared with 81.60\% for the overdamped variant. Since these variants share the same second-order state structure, the results highlight the importance of the \emph{underdamped regime} for state tracking. The rotation comparisons further show that \emph{data dependence matters}: Mamba-3 substantially outperforms variants with fixed or removed rotations across all three tasks. Thus, effective state tracking benefits from both oscillatory dynamics and the ability to adapt phase to incoming tokens.

\begin{findingbox}
\textbf{Main Findings: Oscillation supports effective state tracking when its phase adapts to incoming information.}
Underdamped dynamics outperform critically damped and overdamped alternatives, while data-dependent rotations outperform fixed rotations.
\end{findingbox}




\section{Oscillation in Language Modeling}

\begin{wrapfigure}[18]{r}{0.45\linewidth}
    \vspace{-15pt}
    \centering
    \includegraphics[width=\linewidth]{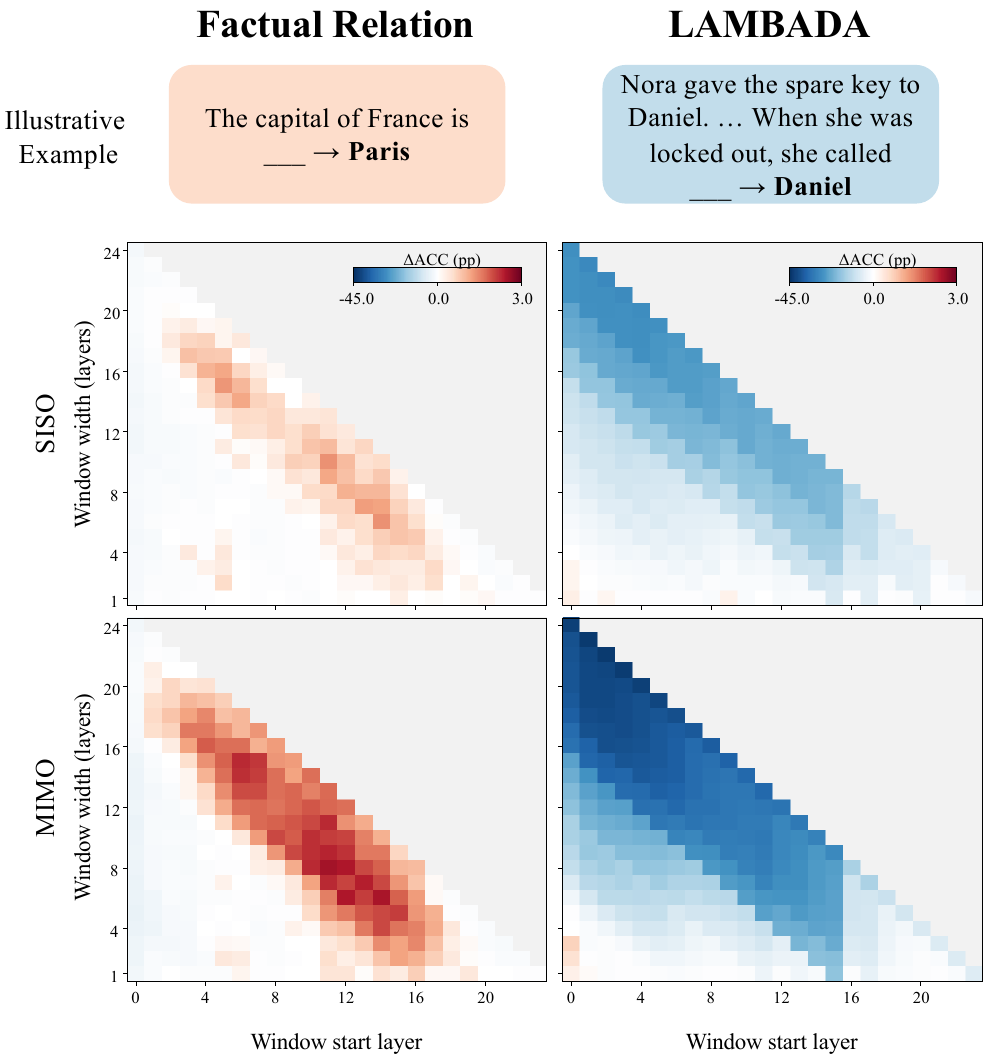}
    \caption{
        Accuracy changes after phase removal in contiguous
        layer windows.
    }
    \label{fig:phase_ablation}
\end{wrapfigure}


We next examine which language tasks depend on oscillatory dynamics, whether oscillation improves long-context performance, and how learned phase shapes recurrent state propagation.

\subsection{Oscillation across Language Tasks}
\label{sec:shortandlong}


\leavevmode\noindent\textbf{Experimental setup.} We compare factual relation prediction on Relation~\citep{hernandez2024linearity} with context-dependent word prediction on LAMBADA~\citep{paperno2016lambada} to assess how reliance on oscillation varies across tasks. Following \citet{sharma2024locating}, we set the rotation phase to zero within consecutive layer windows in pretrained 1.5B Mamba-3 SISO and MIMO models. We vary the window position and size and measure accuracy changes relative to the unmodified models. 


\noindent\textbf{Results.} Removing phase rotation has a much stronger effect on LAMBADA than on Relation (Fig.~\ref{fig:phase_ablation}). Relation accuracy changes only modestly, with occasional improvements when rotation is removed from middle layers. LAMBADA accuracy, however, drops substantially across many window positions, and removing rotation from more layers generally increases the degradation. This task-dependent sensitivity suggests that learned phase is particularly important for using contextual information, motivating a closer examination of its role in long-context language modeling.

\finishwrap  

\begin{findingbox}
\textbf{Main Findings: Learned oscillations are particularly important for context-dependent prediction.}
Removing phase rotation has only a modest effect on factual relation prediction but substantially degrades performance on LAMBADA, which requires integrating information across the preceding context.
\end{findingbox}


\subsection{Pretraining and Long-Context Evaluation}
\label{sec:pretrain}

\begin{wrapfigure}[20]{r}{0.4\textwidth}
    \centering
    \vspace{-28pt}
    \includegraphics[width=\linewidth]{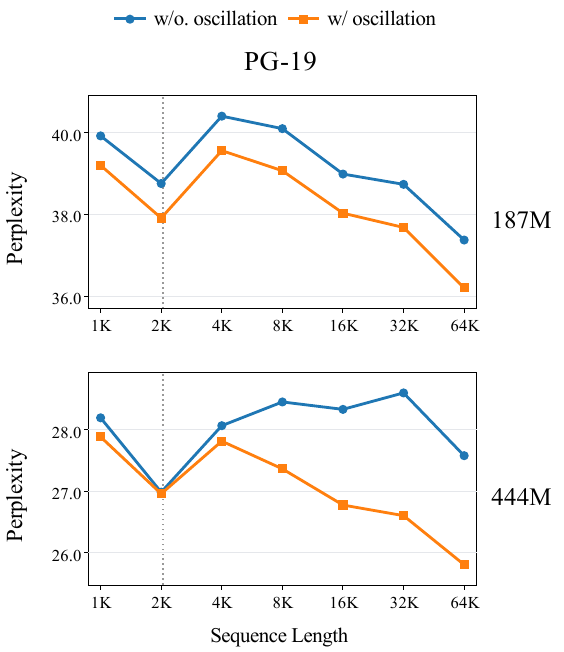}
    \caption{
        PG-19 perplexity across context lengths for models with/without oscillation. The dotted line marks the 2K training context length.
    }
    \label{fig:pg19_ppl}
\end{wrapfigure}


\leavevmode\noindent\textbf{Experimental setup.} We test whether the importance of phase in pretrained models translates into better long-context performance when models are trained from scratch. We train Mamba-3 variants with/without oscillations at the scale of 187M and 444M, each on 100B tokens of FineWeb-Edu dataset~\citep{penedo2024fineweb} with a 2K context length, following \citet{lahoti2026mamba}. We evaluate length generalization through PG-19 perplexity~\citep{rae2019compressive} at context lengths up to 64K and downstream tasks on LongBench-E~\citep{bai2024longbench}. 

\noindent\textbf{Results.} As shown in Fig.~\ref{fig:pg19_ppl}, oscillatory models achieve lower PG-19 perplexity at both scales, with a more pronounced advantage at longer contexts for the larger model. We find that this benefit is not specific to PG-19: following \citet{grazzi2025unlocking}, we further evaluate perplexity on CodeParrot~\citep{tunstall2022natural}, MATH-Hard~\citep{hendrycksmath2021}, TriviaQA~\citep{joshi2017triviaqa}, and SlimPajama~\citep{cerebras2023slimpajama}, finding lower perplexity beyond the training context on most datasets (Appendix~\ref{app:results:generalization}).

On LongBench-E, the variants perform comparably at 187M, scoring 9.53 with oscillation and 9.76 without. At 444M, oscillation increases the average score from 10.61 to \textbf{13.22} (Table~\ref{tab:longbench_e}). These results distinguish the improvements in language-modeling perplexity observed at both scales from the downstream gains observed at 444M.
\begin{table*}[h]
    \centering
    \caption{
        LongBench-E results for models with and without oscillatory dynamics. We compare the non-oscillatory variant and the default oscillatory configuration at two model scales.
    }
    \label{tab:longbench_e}
    \small
    \setlength{\tabcolsep}{3.2pt}
    \renewcommand{\arraystretch}{1.05}

    \resizebox{\textwidth}{!}{%
    \begin{tabular}{
        ll
        cc
        cc
        cc
        cc
        ccc
        cc
        c
    }
        \toprule

        \multirow{2}{*}{\textbf{Params}}
        & \multicolumn{1}{c}{
            \multirow{2}{*}{\textbf{Dynamics}}
        }
        & \multicolumn{2}{c}{\textbf{Synthetic}}
        & \multicolumn{2}{c}{\textbf{Summary}}
        & \multicolumn{2}{c}{\textbf{Single-doc QA}}
        & \multicolumn{2}{c}{\textbf{Multi-doc QA}}
        & \multicolumn{3}{c}{\textbf{Few-shot Learning}}
        & \multicolumn{2}{c}{\textbf{Coding}}
        & \multirow{2}{*}{\textbf{Avg.}}
        \\

        \cmidrule(lr){3-4}
        \cmidrule(lr){5-6}
        \cmidrule(lr){7-8}
        \cmidrule(lr){9-10}
        \cmidrule(lr){11-13}
        \cmidrule(lr){14-15}

        &
        & \textbf{PC}
        & \textbf{PR}
        & \textbf{GR}
        & \textbf{MN}
        & \textbf{MQA}
        & \textbf{QA}
        & \textbf{2WM}
        & \textbf{HQA}
        & \textbf{SS}
        & \textbf{TR}
        & \textbf{TQA}
        & \textbf{LCC}
        & \textbf{RB}
        & \\
        
        \midrule

        \multicolumn{1}{c}{\multirow{2}{*}{187M}}
        & w/o oscillation
        & \textbf{5.96}
        & 6.07
        & 7.75
        & 9.17
        & 9.62
        & \textbf{4.37}
        & \textbf{6.71}
        & \textbf{5.24}
        & 7.87
        & \textbf{12.67}
        & 10.63
        & \textbf{20.25}
        & \textbf{20.57}
        & \textbf{9.76}
        \\

        & w/ oscillation
        & 5.56
        & \textbf{6.64}
        & \textbf{8.93}
        & \textbf{9.25}
        & \textbf{10.02}
        & 4.13
        & 6.25
        & 4.34
        & \textbf{9.02}
        & 8.67
        & \textbf{12.47}
        & 18.32
        & 20.36
        & 9.53
        \\

        \midrule

        \multicolumn{1}{c}{\multirow{2}{*}{444M}}
        & w/o oscillation
        & \textbf{5.96}
        & \textbf{7.19}
        & 8.37
        & \textbf{10.52}
        & 10.71
        & \textbf{5.28}
        & 7.29
        & 5.13
        & 10.36
        & 11.00
        & 19.29
        & 19.58
        & 17.19
        & 10.61
        \\

        & w/ oscillation
        & 3.57
        & 5.95
        & \textbf{8.50}
        & 10.23
        & \textbf{11.38}
        & 5.06
        & \textbf{8.79}
        & \textbf{5.95}
        & \textbf{19.61}
        & \textbf{18.33}
        & \textbf{21.56}
        & \textbf{28.58}
        & \textbf{24.31}
        & \textbf{13.22}
        \\

        \bottomrule
    \end{tabular}%
    }
\end{table*}
Standard benchmarks provide a complementary check on general performance (Table~\ref{app:standard_eval}). Evaluated with the lm-evaluation-harness, oscillatory models achieve lower FineWeb-Edu and LAMBADA perplexity at both scales. Average downstream accuracy is comparable at 187M parameters (43.43 vs.\ 43.63) and modestly higher at 444M parameters (51.08 vs.\ 50.18). Together, these evaluations show that oscillation benefits length generalization while maintaining general performance, with improvements on both long-context and standard downstream tasks at the larger scale.

\begin{table*}[h]
    \centering
    \caption{
        Standard language modeling and downstream evaluation after
        pretraining on 100B tokens of FineWeb-Edu.
        We compare Mamba-3 models with/without oscillatory dynamics
        at two scales. 
    }
    \label{app:standard_eval}
    \small
    \setlength{\tabcolsep}{3.8pt}
    \renewcommand{\arraystretch}{1.00}

    \resizebox{\textwidth}{!}{%
    \begin{tabular}{llcccccccccc}
        \toprule
        \multirow{2}{*}{\textbf{Params}}
        & \multicolumn{1}{c}{
            \multirow{2}{*}{\textbf{Dynamics}}
        }
        & \textbf{FW-Edu}
        & \textbf{LAMB.}
        & \textbf{LAMB.}
        & \textbf{HellaS.}
        & \textbf{PIQA}
        & \textbf{Arc-E}
        & \textbf{Arc-C}
        & \textbf{WinoGr.}
        & \textbf{OBQA}
        & \textbf{Average}
        \\

        \cmidrule(lr){3-12}

        &
        & \textbf{ppl}$\downarrow$
        & \textbf{ppl}$\downarrow$
        & \textbf{acc}$\uparrow$
        & \textbf{acc\_n}$\uparrow$
        & \textbf{acc}$\uparrow$
        & \textbf{acc}$\uparrow$
        & \textbf{acc\_n}$\uparrow$
        & \textbf{acc}$\uparrow$
        & \textbf{acc}$\uparrow$
        & \textbf{acc}$\uparrow$
        \\

        \midrule

        \multicolumn{1}{c}{\multirow{2}{*}{187M}}
        & w/o oscillation
        & 16.32 & 38.04 & 31.73 & 40.86
        & \textbf{67.41} & 60.73
        & \textbf{28.84} & \textbf{53.83}
        & 22.00 & \textbf{43.63}
        \\

        & w/ oscillation
        & \textbf{16.18} & \textbf{32.97}
        & \textbf{33.59} & \textbf{41.24}
        & 67.03 & \textbf{61.28}
        & 25.85 & 51.62 & \textbf{23.40} & 43.43
        \\

        \midrule

        \multicolumn{1}{c}{\multirow{2}{*}{444M}}
        & w/o oscillation
        & 12.68 & 18.38 & 41.01 & 52.40
        & 70.62 & 69.02 & 32.94
        & \textbf{57.06} & \textbf{28.20} & 50.18
        \\

        & w/ oscillation
        & \textbf{12.60} & \textbf{17.19}
        & \textbf{42.83} & \textbf{52.65}
        & \textbf{71.65} & \textbf{70.45}
        & \textbf{35.15} & 56.83 & 28.00
        & \textbf{51.08}
        \\

        \bottomrule
    \end{tabular}%
    }
\end{table*}

\begin{findingbox}
\textbf{Main Findings: Oscillation improves length generalization, with greater downstream benefits at the larger model scale.}
Models with oscillation achieve lower long-context perplexity at both scales. Downstream scores improve at 444M and remain comparable at 187M.
\end{findingbox}

\Needspace{5\baselineskip}
\subsection{The Role of Phase in Long-context Language Modeling}
\label{eval}

\begin{wrapfigure}[13]{R}{0.5\textwidth}
    \centering
    \vspace{-16pt}
    \includegraphics[width=\linewidth]{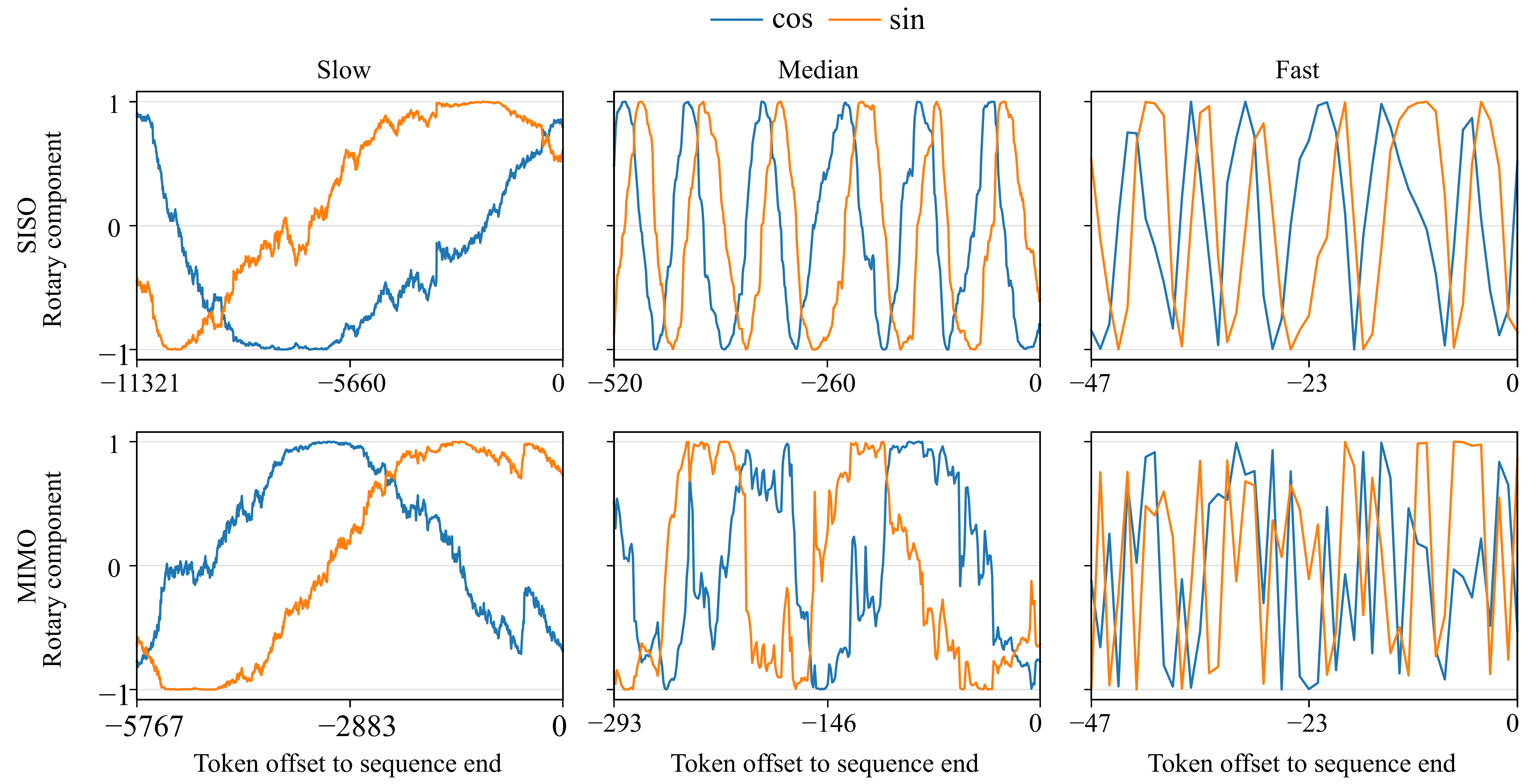}
    \caption{
        Representative phase-channel trajectories in pretrained Mamba-3, showing oscillations at various timescales during language processing.
    }
    \label{fig:wave}
\end{wrapfigure}

To understand the long-context benefits, we investigate how learned phase shapes recurrent computation in pretrained 1.5B Mamba-3. We examine phase trajectories and use interventions to study their contribution to prediction, state accumulation, and write--read alignment.

\noindent\textbf{Learned Phase Dynamics.} Fig.~\ref{fig:wave} shows representative phase channels during natural language processing. Their trajectories exhibit oscillations at different timescales, reflecting the temporal structure of the learned rotations. 

\noindent\textbf{Interventions on Phase Structure.}
We perturb the rotation angles at each recurrent step through three interventions: \emph{phase shuffle} permutes angles across rotary state pairs, \emph{phase removal} sets them to zero, and \emph{phase reversal} flips their signs. These interventions leave decay factors unchanged. For comparison, we apply \emph{decay permutation}, which permutes decay factors across hidden states while preserving phases. We evaluate all interventions in SISO and MIMO models and give implementation details in Appendix~\ref{app:eval}.

As shown in Table~\ref{tab:phase-pg19}, all three phase interventions substantially increase PG-19 perplexity. In SISO, their effects grow strongly with context length: phase removal and reversal raise perplexity to $41.33\times$ and $74.10\times$ the native value at 64K. Phase shuffle is less damaging than decay permutation at 2K but more damaging at 64K. In MIMO, phase interventions have weaker length dependence but consistently cause greater degradation than decay permutation. These results show that long-context prediction heavily relies on the learned phase structure and the assignment to state pairs.

\begin{table}[htbp]
    \centering
    \caption{
        PG-19 perplexity under phase interventions. Higher indicates more damaging.
    }
    \label{tab:phase-pg19}
    \small
    \setlength{\tabcolsep}{5pt}
    \renewcommand{\arraystretch}{1.1}

    \begin{tabular}{@{}clccc@{}}
        \toprule
        Model & Intervention & 2K & 16K & 64K \\
        \midrule

        \multirow{5}{*}{SISO}
        & Native
        & 21.09
        & 21.41
        & 20.60 \\

        & Decay permutation
        & 114.52 $(5.43\times)$
        & 126.51 $(5.91\times)$
        & 125.27 $(6.08\times)$ \\

        & Phase shuffle
        & 50.24 $(2.38\times)$
        & 96.05 $(4.49\times)$
        & 149.37 $(7.25\times)$ \\

        & Phase removal
        & 67.99 $(3.22\times)$
        & 285.66 $(13.34\times)$
        & 851.49 $(41.33\times)$ \\

        & Phase reversal
        & \textbf{152.27} $\boldsymbol{(7.22\times)}$
        & \textbf{597.91} $\boldsymbol{(27.93\times)}$
        & \textbf{1526.52} $\boldsymbol{(74.10\times)}$ \\

        \midrule

        \multirow{5}{*}{MIMO}
        & Native
        & 21.12
        & 21.47
        & 20.52 \\

        & Decay permutation
        & 25.87 $(1.22\times)$
        & 25.94 $(1.21\times)$
        & 24.89 $(1.21\times)$ \\

        & Phase shuffle
        & 64.43 $(3.05\times)$
        & 70.44 $(3.28\times)$
        & 65.72 $(3.20\times)$ \\

        & Phase removal
        & \textbf{197.17} $\boldsymbol{(9.34\times)}$
        & \textbf{320.40} $\boldsymbol{(14.92\times)}$
        & \textbf{309.36} $\boldsymbol{(15.08\times)}$ \\

        & Phase reversal
        & 74.34 $(3.52\times)$
        & 78.76 $(3.67\times)$
        & 73.89 $(3.60\times)$ \\

        \bottomrule
    \end{tabular}
\end{table}

\begin{wraptable}{R}{0.45\textwidth}
    \centering
    \small
    \vspace{-12pt}
    \setlength{\tabcolsep}{4pt}
    \caption{
        Hidden-state statistics under native phase and
        phase removal on PG-19.
    }
    \label{tab:phase-accumulation}

    \resizebox{\linewidth}{!}{%
    \begin{tabular}{@{}llrrrr@{}}
        \toprule
        \multirow{2}{*}{Model}
        & \multirow{2}{*}{Length}
        & \multicolumn{2}{c}{Cosine Similarity}
        & \multicolumn{2}{c}{RMS Norm} \\
        \cmidrule(lr){3-4}
        \cmidrule(lr){5-6}
        &
        & Native & Remove
        & Native & Remove \\
        \midrule

        \multirow{2}{*}{SISO}
        & 16K & 0.293 & 0.429 & 7.635 & 8.899 \\
        & 64K & 0.294 & 0.474 & 7.621 & 9.437 \\

        \midrule

        \multirow{2}{*}{MIMO}
        & 16K & 0.241 & 0.488 & 10.073 & 12.022 \\
        & 64K & 0.242 & 0.499 & 10.078 & 12.047 \\

        \bottomrule
    \end{tabular}%
    }
\end{wraptable}

\noindent\textbf{Phase and Recurrent State Accumulation.} We examine how phase affects the combination of retained history and incoming updates using two statistics: their cosine similarity and the RMS norm of the recurrent state. Cosine similarity measures directional alignment, with larger values favoring constructive accumulation, while the RMS norm measures the resulting state magnitude. 

As shown in Table~\ref{tab:phase-accumulation}, removing phase increases both history--update alignment and state magnitude at 16K and 64K in SISO and MIMO models. These observations suggest that learned rotations moderate constructive accumulation by reorienting retained information relative to new updates. Previous work links excessive state-norm growth to poor length extrapolation in Mamba-based models~\citep{lu2025mamba,shen2026collapse}, making this regulation a possible contributor to the long-context benefits of oscillation. Following \citet{shen2026collapse}, we further analyze how rotation affects state accumulation in Appendix~\ref{app:math:accumulation}.

Notice that the effect of removing phase from pretrained models differs from training without oscillation. Our non-oscillatory models trained from scratch also maintain stable perplexity beyond the training context, without the severe degradation (Sec.~\ref{sec:pretrain}). Oscillation therefore improves long-context performance in our experiments but is not required for stable length extrapolation. Understanding how models trained without oscillation achieve this stability remains an open question.

\finishwrap

\begin{wraptable}[9]{r}{0.45\textwidth}
    \centering
    \small
    \setlength{\tabcolsep}{4pt}
    \vspace{-8pt}
    \caption{
        PG-19 perplexity under phase removal during the final 100 tokens.
    }
    \label{tab:bc-phase-tail}
    \resizebox{\linewidth}{!}{%
    \begin{tabular}{@{}lcc@{}}
        \toprule
        Intervention & SISO & MIMO \\
        \midrule
        Native
        & 20.60
        & 20.52 \\
        Remove $B$
        & 186.28 $(9.04\times)$
        & 100.41 $(4.89\times)$ \\
        Remove $C$
        & 250.90 $(12.18\times)$
        & 117.68 $(5.74\times)$ \\
        Remove both
        & 42.79 $(2.08\times)$
        & 59.46 $(2.90\times)$ \\
        \bottomrule
    \end{tabular}%
    }
\end{wraptable}

\noindent\textbf{Phase Consistency Between Writing and Readout.} We test whether writing and readout depend on compatible phase coordinates by removing phase rotations from $B$ (writing), $C$ (readout), or both. Interventions apply across all layers during the final 100 tokens of 64K windows, while the preceding context is processed with native phase.

Table~\ref{tab:bc-phase-tail} shows that removing rotation from either side alone substantially increases perplexity, whereas removing it from both sides partially restores performance. This recovery supports the importance of phase consistency between writing and readout. The remaining gap between joint removal and the native model may also reflect a mismatch with the state accumulated under native rotations during the preceding context.

\finishwrap

\begin{findingbox}
\textbf{Main Findings: Learned phase regulates how information accumulates and is accessed over long contexts.}
Disrupting phase degrades prediction, while removing rotation increases alignment and accumulation of old and new information. Consistent phase between writing and readout also matters for accurate prediction.
\end{findingbox}

\section{Conclusion}
\label{sec:conclusion}
We interpret Mamba-3's data-dependent rotations as underdamped second-order dynamics, connecting its recurrent structure to oscillatory computation. Controlled experiments show that underdamping benefits state tracking and that oscillation improves long-context prediction and length generalization. Phase interventions reveal the importance of learned rotations for long-context prediction. Hidden-state measurements suggest that rotation moderates information accumulation, while write--read interventions highlight the importance of phase consistency. These findings highlight complementary roles for decay and phase: decay controls how much information persists, while phase shapes how it combines with new inputs and is accessed over time.

\noindent\textbf{Connections to Linear Attention and Recurrent Architectures.} This dynamical perspective links SSMs, linear-attention models, and other recurrent architectures through their state evolution, motivating input-dependent oscillation as a design dimension for future sequence models (Appendix~\ref{app:extension}). We further discuss state dynamics as a design principle in our blog post, \emph{Back to Dynamics}: \url{https://shutian-liang.github.io/backtodynamics/}.

\clearpage
\subsection*{AI Use Statement}

In this work, we used generative AI tools for language polishing and code generation. All AI-assisted text was reviewed and revised by the authors. AI-generated code was inspected, tested, and verified by the authors for correctness before being used in the experiments. We take full responsibility for the final content of this work, including all text, claims, code, and other artifacts produced with the aid of generative AI.

\bibliographystyle{references}
\begingroup
\sloppy
\bibliography{references}
\endgroup

\clearpage
\appendix

\startcontents[appendices]

\section*{Appendix Contents}
\printcontents[appendices]{}{1}{}

\clearpage
\section{Extension to Linear Attention and Recurrent Models}
\label{app:extension}

Although this work primarily focuses on Mamba-3, the dynamical-systems perspective also extends to the homogeneous transitions of other linear-attention models. Table~\ref{tab:deltanet} summarizes representative update rules and their homogeneous dynamics in an interval. Complex KDA can admit oscillatory modes, while Momentum DeltaNet (MDN) explicitly introduces a momentum state that permits damped oscillations~\citep{siems2026complexkdaunderstandingenhancing,huang2026mdn}. \textbf{How to guarantee and control underdamped modes within linear-attention updates remains an interesting direction for future work.} Below, we characterize their spectral regimes and derive continuous-time generators where they exist.
\begin{table}[h]
    \centering
    \caption{
        Linear attention models and their homogeneous dynamics within an interval. Delta-rule models use unit-norm keys.
        For DeltaNet, Gated DeltaNet, and KDA, we assume $\beta_t\in(0,1)$; the decay gates in Gated DeltaNet and KDA lie in $(0,1)$. Complex KDA allows $\beta_t \in [0,2]$ and $\alpha_{t,i} \in [-1,1]$. MDN uses its constrained gate parameterization.
    }
    \label{tab:deltanet}
    \small
    \setlength{\tabcolsep}{3pt}
    \renewcommand{\arraystretch}{1.15}
    \begin{tabular}{@{}lll@{}}
        \toprule
        Model & Update rule & Dynamics \\
        \midrule
        Linear Attention~\citep{pmlr-v119-katharopoulos20a}
        & $S_t = S_{t-1} + k_t v_t^\top$
        & No Decay \\

        DeltaNet~\citep{schlag2021linear,yang2024parallelizing}
        & $S_t = H_t S_{t-1} + \beta_t k_t v_t^\top $
        & Decay with a preserved mode \\

        Gated DeltaNet~\citep{yang2025gated}
        & $S_t = \alpha_t H_t S_{t-1}
                 + \beta_t k_t v_t^\top$
        & Non-oscillatory two-rate decay\\
        KDA~\citep{kimiteam2025kimilinearexpressiveefficient,team2026kimi}
        & $S_t = H_t D_t S_{t-1}
                 + \beta_t k_t v_t^\top$
        & Non-oscillatory modal decay \\
        
        Complex KDA~\citep{siems2026complexkdaunderstandingenhancing}
        & $S_t = H_t D_t S_{t-1}
                 + \beta_t k_t v_t^\top$
        & Can admit oscillatory modes \\

        MDN~\citep{huang2026mdn}
        & $\begin{aligned}
            M_t &= \mu_tM_{t-1}-\eta_tk_te_t^\top,\\
            S_t &= \alpha_tS_{t-1}-\beta_tM_t
          \end{aligned}$
        & Can admit oscillatory modes \\

        \bottomrule
    \end{tabular}
\end{table}
\begin{table}[h]
    \centering
    \caption{Notation and definitions.}
    \label{tab:notations}
    \small
    \begin{tabular}{ll}
        \toprule
        Notation & Meaning \\
        \midrule
        $I$ & Identity matrix \\
        $S_t$ & Recurrent state matrix \\
        $q_t$ & Query vector \\
        $k_t$ & Key vector \\
        $v_t$ & Value vector \\
        $o_t$ & Readout vector, $S_t^\top q_t$ \\
        $T_t$ & State transition matrix \\
        $U_t$ & Additive input term \\
        $\beta_t$ & Input-dependent delta-rule update gate \\
        $H_t$ & Delta-rule transition matrix,
                  $I-\beta_t k_t k_t^\top$ \\
        $\alpha_t$ & Scalar decay gate \\
        $\boldsymbol{\alpha}_t$ & Vector of per-channel gates \\
        $D_t$ & Diagonal gate matrix,
                  $\operatorname{Diag}(\boldsymbol{\alpha}_t)$ \\
        $M_t$ & Momentum state matrix in MDN \\
        $\mu_t$ & Momentum retention gate in MDN \\
        $\eta_t$ & Correction scale in MDN \\
        $e_t$ & Prediction error, 
        $v_t-\alpha_tS_{t-1}^{\top}k_t$ \\
        \bottomrule
    \end{tabular}
\end{table}
\subsection{Homogeneous state propagation}

Following Sec.~\ref{sec:second_order_rope}, we analyze homogeneous state propagation with coefficients held fixed within each interval. For the matrix-valued recurrence
\begin{equation}
S_t=T_tS_{t-1}+U_t,
\qquad
o_t=S_t^\top q_t,
\label{eq:extension_recurrence}
\end{equation}
we omit the additive term $U_t$ and examine the transition $T_t$, which acts on each column
$s_{t-1}\in\mathbb{R}^{d_k}$ of $S_{t-1}$. For MDN, we instead analyze the block transition of the augmented state $(S_t,M_t)$. Within an interval of length $\Delta>0$, we suppress the index $t$ and construct continuous-time realizations of the frozen transitions.

For each selected invariant two-dimensional subsystem of the
single-state models, we use $x=(u,v)^\top$ to denote its two
scalar state coordinates. We specify these coordinates for each
transition below and, when they are coupled, eliminate $v$ from
the corresponding first-order dynamics to obtain a second-order
equation for $u$. Where the resulting equation has
$\omega_0>0$, we compare it with the canonical form
\begin{equation}
\ddot u
+2\zeta\omega_0\dot u
+\omega_0^2u
=0.
\label{eq:extension_canonical}
\end{equation}
Modes with a zero continuous-time eigenvalue, such as DeltaNet's preserved mode, are described directly rather than assigned a finite damping ratio. MDN has an additional momentum state, so we analyze its augmented state $(S_t,M_t)$ separately below. Here $u$ and $v$ are scalar coordinates of the selected subsystem; they need not be entries of the original state column $s$. All derivatives are taken within the fixed interval.

\subsection{Linear Attention}
Linear Attention has $T=I$. Its homogeneous update is reproduced by $\dot S=0$, so the state is preserved within the interval. The full recurrence accumulates input contributions,
\begin{equation}
    S_t=S_0+\sum_{j=1}^{t}U_{j}.
\end{equation}
\subsection{DeltaNet}

Let $k\in\mathbb{R}^{d_k}$ be a unit-norm key, with $d_k\geq2$, and define
\begin{equation}
\Pi=kk^\top,
\qquad
\Pi^2=\Pi.
\end{equation}
For $0<\beta<1$, the homogeneous transition is
\begin{equation}
T=I-\beta\Pi
=(I-\Pi)+(1-\beta)\Pi.
\end{equation}
Thus, the component along $k$ is multiplied by $1-\beta$,
while every component orthogonal to $k$ is preserved. Define
\begin{equation}
b=-\frac{\log(1-\beta)}{\Delta}>0,
\qquad
A=-b\Pi.
\label{eq:extension_delta_generator}
\end{equation}
Since $\Pi$ is a projector, $\Pi^n=\Pi$ for every integer $n\geq1$. Expanding the matrix exponential gives
\begin{equation}
\begin{aligned}
e^{A\Delta}
&=\sum_{n=0}^{\infty}
\frac{(A\Delta)^n}{n!} \\
&=I+
\left(
\sum_{n=1}^{\infty}
\frac{(-b\Delta)^n}{n!}
\right)\Pi \\
&=I+\left(e^{-b\Delta}-1\right)\Pi \\
&=I-\beta\Pi
=T.
\end{aligned}
\end{equation}
Hence $\dot s=As$ reproduces the discrete homogeneous
transition over the interval. Choose an orthonormal basis
$\{k,w_1,\ldots,w_{d_k-1}\}$ satisfying
\begin{equation}
k^\top w_i=0,
\qquad
w_i^\top w_j=\delta_{ij}.
\end{equation}
The state admits the decomposition
\begin{equation}
s
=
\mu k
+
\sum_{i=1}^{d_k-1}\nu_iw_i,
\qquad
\mu=k^\top s,
\qquad
\nu_i=w_i^\top s.
\end{equation}
These scalar coordinates are projections onto the chosen basis
directions rather than necessarily entries of $s$ in its original
coordinate system. Using $A=-bkk^\top$, together with $k^\top k=1$ and $k^\top w_i=0$, we have
\begin{equation}
\begin{aligned}
Ak
&= -bkk^\top k
 = -bk(k^\top k)
 = -bk, \\
Aw_i
&= -bkk^\top w_i
 = -bk(k^\top w_i)
 = 0.
\end{aligned}
\end{equation}
Thus, each plane $\mathcal E_i = \operatorname{span}\{k,w_i\}$ is invariant. On this plane, the modal coordinates satisfy
\begin{equation}
\dot\mu = k^\top\dot s = k^\top As = -bk^\top kk^\top s =
-b(k^\top k)(k^\top s)= -b\mu,
\end{equation}
Similarly,
\begin{equation}
\dot\nu_i
=
w_i^\top\dot s
=
w_i^\top As
=
-bw_i^\top kk^\top s
=
-b(w_i^\top k)(k^\top s)
=
0,
\end{equation}
To obtain an equivalent second-order representation on $\mathcal E_i$, select one
plane $\mathcal E_i$ and introduce the invertible coordinates
\begin{equation}
u=\mu+\nu_i,
\quad
v=\mu,
\quad
x=(u,v)^{\top},
\end{equation}
so $(u,v)$ retains the full two-dimensional state on
$\mathcal E_i$. Their dynamics are
\begin{equation}
\dot u
=
\dot\mu+\dot\nu_i
=
-bv,
\quad
\dot v
=
-bv,
\end{equation}
Hence
\begin{equation}
\ddot u =-b\dot v =b^2v =-b\dot u.
\end{equation}
Therefore,
\begin{equation}
\ddot u+b\dot u=0.
\label{eq:extension_delta_second_order}
\end{equation}
On each plane $\mathcal E_i$, the eigenvalues are $-b$ and $0$, corresponding to a decaying and a preserved mode, respectively. In the full state space, however, $-b$ has multiplicity one, while $0$ has multiplicity $d_k-1$. Thus, although every orthogonal direction $w_i$ can be represented by a second-order representation, all such planes share the same decaying mode along $k$ (the coordinate $\mu$, or $v$ after the transformation). \textbf{They are overlapping representations of the full system, rather than independent two-dimensional subsystems. The same shared-direction structure persists in Gated DeltaNet.
}

\subsection{Gated DeltaNet}

Gated DeltaNet uses $T=\alpha H$ with $0<\alpha<1$~\citep{yang2025gated}.
With $b$ and $\Pi$ defined above, let
\begin{equation}
    \gamma=-\frac{\log\alpha}{\Delta}>0,
    \quad
    A=-\gamma I-b\Pi.
\end{equation}
Since $I$ and $\Pi$ commute,
\begin{equation}
    \begin{aligned}
        e^{A\Delta}
        &=e^{-\gamma\Delta}
          \bigl[I+(e^{-b\Delta}-1)\Pi\bigr]\\
        &=\alpha(I-\beta\Pi)
         =\alpha H
         =T.
    \end{aligned}
    \label{eq:extension_gated_delta_generator}
\end{equation}
Hence $\dot s=As$ reproduces the discrete homogeneous transition over the interval.

Use the orthonormal basis $\{k,w_1,\ldots,w_{d_k-1}\}$ defined in the DeltaNet subsection. The action of $A$ on its basis vectors is
\begin{equation}
    \begin{aligned}
        Ak
        &=-\gamma k-bk(k^\top k)
         =-(\gamma+b)k,\\
        Aw_i
        &=-\gamma w_i-bk(k^\top w_i)
         =-\gamma w_i.
    \end{aligned}
\end{equation}
Thus each plane $\mathcal E_i=\operatorname{span}\{k,w_i\}$ is invariant. Its modal coordinates $\mu=k^\top s$ and $\nu_i=w_i^\top s$ satisfy
\begin{equation}
    \begin{aligned}
        \dot\mu
        &=k^\top As
         =-(\gamma+b)\mu,\\
        \dot\nu_i
        &=w_i^\top As
         =-\gamma\nu_i.
    \end{aligned}
\end{equation}
The key-direction coordinate $\mu$ decays at rate $\gamma+b$, while every orthogonal coordinate $\nu_i$ decays at rate $\gamma$.

As in DeltaNet, select one plane $\mathcal E_i$ and define
\begin{equation}
    u=\mu+\nu_i,
    \quad
    v=\mu,
    \quad
    x=(u,v)^\top.
\end{equation}
This coordinate change is invertible on $\mathcal E_i$, with $\mu=v$ and $\nu_i=u-v$. The dynamics become
\begin{equation}
    \begin{aligned}
        \dot u
        &=\dot\mu+\dot\nu_i 
        =-(\gamma+b)v-\gamma(u-v)
        =-\gamma u-bv,\\
        \dot v
        &=-(\gamma+b)v.
    \end{aligned}
\end{equation}
Differentiating $\dot u$ and using
$bv=-(\dot u+\gamma u)$ gives
\begin{equation}
    \begin{aligned}
        \ddot u
        &=-\gamma\dot u-b\dot v\\
        &=-\gamma\dot u+b(\gamma+b)v\\
        &=-\gamma\dot u
          -(\gamma+b)(\dot u+\gamma u)\\
        &=-(2\gamma+b)\dot u-\gamma(\gamma+b)u.
    \end{aligned}
\end{equation}
Therefore,
\begin{equation}
    \ddot u+(2\gamma+b)\dot u
    +\gamma(\gamma+b)u=0.
    \label{eq:extension_gated_delta_second_order}
\end{equation}
The canonical parameters are
\begin{equation}
    \omega_0=\sqrt{\gamma(\gamma+b)},
    \quad
    \zeta=
    \frac{2\gamma+b}{2\sqrt{\gamma(\gamma+b)}}>1.
\end{equation}
On each plane $\mathcal E_i$, the characteristic roots are $-(\gamma+b)$ and $-\gamma$, so the subsystem is overdamped. In the full state space, $-(\gamma+b)$ has multiplicity one, while $-\gamma$ has multiplicity $d_k-1$. As in DeltaNet, each $\nu_i$ can be paired with $\mu$, \textbf{but all such planes share the single key-direction mode}. Their second-order representations therefore overlap.
\subsection{Kimi Delta Attention}

KDA replaces the scalar gate with a diagonal gate,
$T=HD$, where
$D=\operatorname{Diag}(\boldsymbol{\alpha})$
\citep{kimiteam2025kimilinearexpressiveefficient,team2026kimi}.
Assume $0<\beta<1$ and $0<\alpha_i<1$.
Although $T=HD$ need not be symmetric, the similarity
transform
\begin{equation}
    \widehat T
    =D^{1/2}TD^{-1/2}
    =D^{1/2}HD^{1/2}
    \label{eq:extension_kda_similarity}
\end{equation}
is symmetric. Since
$(1-\beta)I\preceq H\preceq I$, it satisfies
\begin{equation}
    0\prec(1-\beta)D
    \preceq\widehat T
    \preceq D
    \prec I.
\end{equation}
Therefore $T$ has a real eigenbasis and eigenvalues
$\lambda_\ell\in(0,1)$. Write
\begin{equation}
    T
    =Q\operatorname{Diag}
      (\lambda_1,\ldots,\lambda_{d_k})Q^{-1},
\end{equation}
where the columns $q_\ell$ of $Q$ are eigenvectors
of $T$. For $\Delta>0$, define
\begin{equation}
    b_\ell=-\frac{\log\lambda_\ell}{\Delta}>0,
    \quad
    A=Q\operatorname{Diag}
      (-b_1,\ldots,-b_{d_k})Q^{-1}.
\end{equation}
Then
\begin{equation}
    \begin{aligned}
        e^{A\Delta}
        &=Q\operatorname{Diag}
          (e^{-b_1\Delta},\ldots,e^{-b_{d_k}\Delta})
          Q^{-1}=T.
    \end{aligned}
\end{equation}

For a state column $s$, define its modal coordinates by
\begin{equation}
    c=Q^{-1}s,
    \quad
    s=\sum_{\ell=1}^{d_k}c_\ell q_\ell.
\end{equation}
These are expansion coefficients in the eigenvector
basis, which need not be orthogonal. Since $s=Qc$ and $Q$ is fixed within the frozen interval, $\dot s=Q\dot c$. Using $\dot s=As$, we obtain
\begin{equation}
    \dot c
    =Q^{-1}AQc
    =-\operatorname{Diag}(b_1,\ldots,b_{d_k})c.
\end{equation}
Thus $\dot c_\ell=-b_\ell c_\ell$.
The rates $b_\ell$ describe eigenmodes of $T=HD$.

Select two modes with distinct rates $b_i\ne b_j$.
Their eigenvectors span an invariant plane
$\mathcal E_{ij}=\operatorname{span}\{q_i,q_j\}$.
Let
\begin{equation}
    \mu=c_i=(Q^{-1}s)_i,
    \quad
    \nu=c_j=(Q^{-1}s)_j.
\end{equation}
On this plane,
\begin{equation}
    \dot\mu=-b_i\mu,
    \quad
    \dot\nu=-b_j\nu.
\end{equation}
As in the preceding subsections, define
\begin{equation}
    u=\mu+\nu,
    \quad
    v=\mu,
    \quad
    x=(u,v)^\top.
\end{equation}
This coordinate change is invertible on
$\mathcal E_{ij}$, with $\mu=v$ and $\nu=u-v$.
The dynamics become
\begin{equation}
    \begin{gathered}
        \dot u =\dot\mu+\dot\nu =-b_i v-b_j(u-v)=-b_j u-(b_i-b_j)v,\\
        \dot v =-b_i v.
    \end{gathered}
\end{equation}
Differentiating $\dot u$ and using
$(b_i-b_j)v=-(\dot u+b_j u)$ gives
\begin{equation}
    \begin{aligned}
        \ddot u
        &=-b_j\dot u-(b_i-b_j)\dot v\\
        &=-b_j\dot u+b_i(b_i-b_j)v\\
        &=-b_j\dot u-b_i(\dot u+b_j u)\\
        &=-(b_i+b_j)\dot u-b_ib_j u.
    \end{aligned}
\end{equation}
Therefore,
\begin{equation}
    \ddot u+(b_i+b_j)\dot u+b_ib_j u=0.
    \label{eq:extension_kda_second_order}
\end{equation}
Its canonical parameters are
\begin{equation}
    \omega_0=\sqrt{b_ib_j},
    \quad
    \zeta=
    \frac{b_i+b_j}{2\sqrt{b_ib_j}}>1.
\end{equation}
The characteristic roots are $-b_i$ and $-b_j$; hence, for $b_i\neq b_j$, the subsystem is overdamped. Since KDA can exhibit multiple distinct modal decay rates, disjoint eigenmode pairs can \textbf{form non-overlapping invariant subspaces that evolve independently in modal coordinates.} This differs from DeltaNet and Gated DeltaNet, where all such two-dimensional pairings share the same key-direction mode.
\subsection{Complex KDA}
Complex KDA~\citep{siems2026complexkdaunderstandingenhancing} allows $\beta\in[0,2]$ and $\alpha_i\in[-1,1]$.
Its frozen homogeneous transition is
\begin{equation}
    T=(I-\beta kk^\top)D,
    \quad
    D=\operatorname{Diag}(\boldsymbol{\alpha}),
\end{equation}
To exhibit an oscillatory mode arising from signed channel-wise gating, consider the two-dimensional construction in~\citep{siems2026complexkdaunderstandingenhancing}. Let
\begin{equation}
\begin{gathered}
H_k=I-2kk^\top,\quad T_{\alpha,\theta}=H_kD_\alpha,\quad 
D_\alpha=\operatorname{Diag}(\alpha,1),\quad
k=(\cos(\theta/2),\sin(\theta/2))^\top.
\end{gathered}
\end{equation}
As shown in Appendix~\ref{app:math:ckda_derivations}, expanding the product gives
\begin{equation}
T_{\alpha,\theta}=
\begin{bmatrix}
-\alpha\cos\theta & -\sin\theta\\
-\alpha\sin\theta & \cos\theta
\end{bmatrix}.
\end{equation}
Its discriminant is
\begin{equation}
\mathcal D
=
[\operatorname{tr}(T_{\alpha,\theta})]^2
-4\det(T_{\alpha,\theta})
=
(1-\alpha)^2\cos^2\theta+4\alpha.
\end{equation}
A non-real conjugate pair occurs precisely when
\begin{equation}
\alpha<0
\quad\text{and}\quad
|\cos\theta|
<
\frac{2\sqrt{-\alpha}}{1-\alpha}.
\label{eq:ckda_complex_condition}
\end{equation}
Under this condition, the eigenvalues are
\begin{equation}
\lambda_\pm
=
\frac{(1-\alpha)\cos\theta}{2}
\pm
\frac{i}{2}\sqrt{-\mathcal D}.
\end{equation}
Since
$\lambda_+\lambda_-=\det(T_{\alpha,\theta})=-\alpha$,
their common modulus is $\rho=\sqrt{-\alpha}$.
For $-1<\alpha<0$, we have $0<\rho<1$ and may write
\begin{equation}
\lambda_\pm=\rho e^{\pm i\phi},
\qquad
\phi=
\arccos\left(
\frac{(1-\alpha)\cos\theta}{2\rho}
\right)\in(0,\pi).
\end{equation}

For an interval of length $\Delta>0$, define
\begin{equation}
a=-\frac{\log\rho}{\Delta}>0,
\qquad
\Omega=\frac{\phi}{\Delta}>0.
\end{equation}
The real matrix $T_{\alpha,\theta}$ is similar to
$\rho R(\phi)$. In the corresponding real modal coordinates
$x=(u,v)^\top$, a corresponding continuous-time realization is
\begin{equation}
\dot x=
\begin{bmatrix}
-a&-\Omega\\
\Omega&-a
\end{bmatrix}x,
\qquad
e^{(-aI+\Omega J)\Delta}=\rho R(\phi).
\end{equation}
As in the decay--rotation analysis of Mamba-3 in
Sec.~\ref{sec:second_order_rope}, eliminating $v$ yields
\begin{equation}
\ddot u+2a\dot u+(a^2+\Omega^2)u=0,
\end{equation}
with
$\omega_0=\sqrt{a^2+\Omega^2}$ and
$\zeta=a/\omega_0<1$.
Thus, under these conditions, \textbf{this CKDA mode admits an
underdamped second-order realization.}

\subsection{Momentum DeltaNet}
\label{app:momentum}

Momentum DeltaNet (MDN) augments the recurrent state $S_t$
with a momentum state $M_t$ of the same shape
~\citep{huang2026mdn}. Its update is
\begin{equation}
\begin{aligned}
M_t
&=\mu_t M_{t-1}
-\eta_t k_t
 \bigl(v_t-\alpha_t S_{t-1}^{\top}k_t\bigr)^\top,\\
S_t
&=\alpha_t S_{t-1}-\beta_tM_t.
\end{aligned}
\label{eq:extension_mdn_update}
\end{equation}
Freeze the coefficients and a unit-norm key $k$ within an
interval of length $\Delta>0$, and suppress the index $t$.
To study homogeneous propagation, omit the external value
input $v_t$. With $\Pi=kk^\top$, the coupled update becomes
\begin{equation}
\begin{bmatrix}
S_t\\ M_t
\end{bmatrix}
=
\underbrace{
\begin{bmatrix}
\alpha(I-\beta\eta\Pi)&-\beta\mu I\\
\alpha\eta\Pi&\mu I
\end{bmatrix}}_{\mathcal T}
\begin{bmatrix}
S_{t-1}\\ M_{t-1}
\end{bmatrix}.
\label{eq:extension_mdn_block}
\end{equation}

Consider corresponding columns $s$ and $m$ of $S$ and $M$.
Their key-direction coordinates
\begin{equation}
u=k^\top s,
\qquad
p=k^\top m
\end{equation}
form an invariant two-dimensional subsystem. Since
$\Pi k=k$, its transition is
\begin{equation}
\begin{bmatrix}
u_t\\ p_t
\end{bmatrix}
=
B
\begin{bmatrix}
u_{t-1}\\ p_{t-1}
\end{bmatrix},
\qquad
B=
\begin{bmatrix}
\alpha(1-\beta\eta)&-\beta\mu\\
\alpha\eta&\mu
\end{bmatrix}.
\label{eq:extension_mdn_key_block}
\end{equation}
The trace and determinant of $B$ are
\begin{equation}
\tau=\alpha(1-\beta\eta)+\mu,
\qquad
\det(B)=\alpha\mu.
\end{equation}
Hence its characteristic discriminant is
\begin{equation}
\mathcal D=\tau^2-4\alpha\mu.
\end{equation}
When $\mathcal D<0$, the eigenvalues form a non-real
conjugate pair. For $0<\alpha,\mu<1$, define
\begin{equation}
\rho=\sqrt{\alpha\mu}\in(0,1),
\qquad
\phi=\arccos\left(\frac{\tau}{2\rho}\right)\in(0,\pi).
\end{equation}
When $\mathcal D<0$, the eigenvalues are
$\lambda_\pm=\rho e^{\pm i\phi}$.
Set
\begin{equation}
K=\frac{B-\rho\cos\phi\,I}{\rho\sin\phi}.
\end{equation}
The characteristic equation of $B$ gives $K^2=-I$,
and thus
\begin{equation}
B=\rho(\cos\phi\,I+\sin\phi\,K).
\end{equation}
Define
\begin{equation}
a=-\frac{\log\rho}{\Delta}>0,
\qquad
\Omega=\frac{\phi}{\Delta}>0,
\qquad
A=-aI+\Omega K.
\end{equation}
Since $K^2=-I$, we have $e^{A\Delta}=B$.
Moreover, $K$ is similar over the reals to the rotation
generator $J$ in Sec.~\ref{sec:second_order_rope}.
Thus, when $\mathcal D<0$ and $0<\alpha,\mu<1$,
this frozen key-direction subsystem admits an underdamped
second-order realization.

\section{Related Work}


\paragraph{Oscillatory Dynamics and Rotations in Recurrent Models.} Oscillatory systems provide a dynamical foundation for recurrent architectures. CoRNN~\citep{rusch2021coupled} discretizes coupled nonlinear oscillators and analyzes the stability of states and gradients, while Neural Wave Machines~\citep{keller2023neural} use locally coupled oscillators to learn spatiotemporally structured representations. LinOSS~\citep{rusch2025oscillatory} builds state-space models from forced harmonic oscillators, and D-LinOSS~\citep{boyer2025learning} introduces learnable dissipation to control their damping. Selective RoPE~\citep{movahedi2026selective} incorporates input-dependent rotations into selectively decaying linear recurrences, improving state tracking and language modeling. Our work connects Mamba-3 to this family of dynamical models through an underdamped second-order interpretation of its decay--rotation transition. This connection enables controlled comparisons across damping regimes and an analysis of how learned phase shapes recurrent state accumulation.


\paragraph{Complex Parameterizations and Spectral Structure in Recurrent Models.}The spectrum of a recurrent transition governs how states persist and evolve. Early unitary recurrent networks use complex-valued states and norm-preserving transitions to facilitate learning long-term dependencies~\citep{arjovsky2016unitary,wisdom2016full,jing2017tunable}. Complex parameterizations also underpin state-space and linear recurrent architectures such as S4~\citep{gu2021efficiently}, S4D~\citep{gu2022parameterization}, S5~\citep{smith2023simplified}, and LRU~\citep{orvieto2023resurrecting}. Complementing these architectural developments, \citet{pmlr-v235-orvieto24a} study how complex eigenvalues support long-range information retention, and \citet{grazzi2025unlocking} show that allowing negative eigenvalues overcomes limitations of positive-spectrum recurrences on state tracking and improves length generalization. We examine Mamba-3's complex-conjugate modes through their damping and phase dynamics, linking spectral structure to controlled training results and phase interventions in pretrained language models.


\section{Mathematical Derivations}

\subsection{Matrix Exponential of the Rotation Generator}
\label{app:rotation_exponential}

We provide the derivation of the matrix exponential used in
Section~\ref{sec:second_order_rope}. Let
\begin{equation}
    J=
    \begin{bmatrix}
        0 & -1\\
        1 & 0
    \end{bmatrix},
\end{equation}
which satisfies
\begin{equation}
    J^2=-I.
\end{equation}
It follows that
\begin{equation}
    J^{2k}=(-1)^k I,
    \qquad
    J^{2k+1}=(-1)^k J.
\end{equation}

Using the Taylor expansion of the matrix exponential,
\begin{equation}
    e^{\phi J}
    =
    \sum_{k=0}^{\infty}
    \frac{(\phi J)^k}{k!},
\end{equation}
and separating the even and odd powers gives
\begin{align}
    e^{\phi J}
    &=
    \sum_{k=0}^{\infty}
    \frac{\phi^{2k}J^{2k}}{(2k)!}
    +
    \sum_{k=0}^{\infty}
    \frac{\phi^{2k+1}J^{2k+1}}{(2k+1)!} \\
    &=
    \left[
        \sum_{k=0}^{\infty}
        \frac{(-1)^k\phi^{2k}}{(2k)!}
    \right]I
    +
    \left[
        \sum_{k=0}^{\infty}
        \frac{(-1)^k\phi^{2k+1}}{(2k+1)!}
    \right]J \\
    &=
    \cos(\phi)I+\sin(\phi)J.
\end{align}
Substituting the definition of $J$ yields
\begin{equation}
    e^{\phi J}
    =
    \begin{bmatrix}
        \cos\phi & -\sin\phi\\
        \sin\phi &  \cos\phi
    \end{bmatrix}
    =
    R(\phi).
\end{equation}

For the generator
\begin{equation}
    A^{\mathrm{rot}}=-aI+\nu J,
\end{equation}
the matrices $I$ and $J$ commute. Hence,
\begin{align}
    e^{A^{\mathrm{rot}}\Delta}
    &=
    e^{(-aI+\nu J)\Delta} \\
    &=
    e^{-a\Delta I}e^{\nu\Delta J} \\
    &=
    e^{-a\Delta}
    \left[
        \cos(\nu\Delta)I
        +
        \sin(\nu\Delta)J
    \right] \\
    &=
    e^{-a\Delta}R(\nu\Delta).
\end{align}
With
$a=-\log\rho/\Delta$ and $\nu=\theta/\Delta$, this becomes
\begin{equation}
    e^{A^{\mathrm{rot}}\Delta}
    =
    \rho R(\theta),
\end{equation}
which establishes the decay--rotation transition used in the main text.

\subsection{Mode Persistence Across Damping Regimes}
\label{app:mode_persistence}

We compare homogeneous second-order dynamics at a fixed natural
frequency $\omega_0>0$:
\begin{equation}
    \dot s=A(a)s,
    \qquad
    A(a)=
    \begin{bmatrix}
        0 & 1\\
        -\omega_0^2 & -2a
    \end{bmatrix}.
\end{equation}
The eigenvalues are
\begin{equation}
    \lambda_\pm
    =
    -a\pm\sqrt{a^2-\omega_0^2}.
\end{equation}
The discrete-step statements in
Sec.~\ref{sec:eigenmode_memory} follow by evaluating the
continuous-time flow at $t=n\Delta$.

\paragraph{Overdamped dynamics.}
For $a>\omega_0$, let
\begin{equation}
    r=\sqrt{a^2-\omega_0^2}.
\end{equation}
The eigenvalues are
$\lambda_\pm=-a\pm r$, where $\lambda_+$ is the slower mode.
Their relative magnitude satisfies
\begin{equation}
    \lim_{t\to\infty}
    \left|
    \frac{e^{\lambda_-t}}
         {e^{\lambda_+t}}
    \right|
    =
    \lim_{t\to\infty}e^{-2rt}
    =
    0.
\end{equation}
Thus, for initial states with a nonzero slow-mode component,
the fast mode becomes negligible and propagation is
asymptotically dominated by the slow eigendirection.
The dominant exponential decay rate is
\begin{equation}
    a-r
    =
    a-\sqrt{a^2-\omega_0^2}.
\end{equation}

\paragraph{Critically damped dynamics.}
At $a=\omega_0$, write
\begin{equation}
    A=-\omega_0 I+N,
    \qquad
    N=
    \begin{bmatrix}
        \omega_0 & 1\\
        -\omega_0^2 & -\omega_0
    \end{bmatrix}.
\end{equation}
Since $N^2=0$,
\begin{equation}
    e^{At}
    =
    e^{-\omega_0t}(I+tN).
\end{equation}
The polynomial factor does not change the asymptotic exponential
decay rate, which is therefore $\omega_0$. Moreover,
\begin{equation}
    \lim_{t\to\infty}
    \frac{e^{\omega_0t}}{t}e^{At}
    =
    \lim_{t\to\infty}
    \left(
        \frac{I}{t}+N
    \right)
    =
    N.
\end{equation}
Since
\begin{equation}
    N=
    \begin{bmatrix}
        1\\
        -\omega_0
    \end{bmatrix}
    \begin{bmatrix}
        \omega_0 & 1
    \end{bmatrix},
\end{equation}
we have $\operatorname{rank}(N)=1$.
Thus, after removing the common exponential decay and polynomial
factor, the propagation operator becomes asymptotically rank one.

\paragraph{Underdamped dynamics.}
For $0<a<\omega_0$, let
\begin{equation}
    \nu=\sqrt{\omega_0^2-a^2}.
\end{equation}
The eigenvalues are
$\lambda_\pm=-a\pm i\nu$, and hence
\begin{equation}
    \left|
    \frac{e^{\lambda_-t}}
         {e^{\lambda_+t}}
    \right|
    =
    1
    \qquad
    \text{for all }t\ge0.
\end{equation}
Thus, neither mode is preferentially attenuated.
Moreover, there exists an invertible change of coordinates $T$
such that
\begin{equation}
    T A(a)T^{-1}
    =
    -aI+\nu J,
    \qquad
    J=
    \begin{bmatrix}
        0 & -1\\
        1 & 0
    \end{bmatrix}.
\end{equation}
Therefore,
\begin{equation}
    T e^{A(a)t}T^{-1}
    =
    e^{-at}R(\nu t).
\end{equation}
In these rotation coordinates, both state directions decay with
the same radial factor while their relative two-dimensional
geometry is preserved. The exponential decay rate is $a$.

\paragraph{Comparison.}
At fixed $\omega_0$, the underdamped decay rate satisfies
$a<\omega_0$. For the overdamped regime,
\begin{equation}
    a-\sqrt{a^2-\omega_0^2}
    =
    \frac{\omega_0^2}
         {a+\sqrt{a^2-\omega_0^2}}
    <\omega_0,
\end{equation}
whereas the critically damped rate is exactly $\omega_0$.
Critical damping therefore has the fastest asymptotic exponential
attenuation at fixed natural frequency.

More importantly, the regimes differ in their modal geometry:
underdamped propagation preserves two equally decaying modes,
whereas overdamped propagation suppresses the fast mode and
critical propagation becomes asymptotically rank one.

\subsection{State Accumulation Under Persistent Forcing}
\label{app:math:accumulation}

Prior work has identified recurrent-state growth as a failure mode of Mamba
under context-length extrapolation~\citep{shen2026collapse}.
Consider the scalar recurrence
\begin{equation}
    h_t = \rho h_{t-1} + u,
    \qquad 0 \le \rho < 1,
\end{equation}
with $h_0=0$ and constant input $u$. Then
\begin{equation}
    h_t
    =
    \sum_{k=0}^{t-1} \rho^k u
    =
    \frac{1-\rho^t}{1-\rho}u,
\end{equation}
and hence
\begin{equation}
    h_\infty = \frac{1}{1-\rho}u.
\end{equation}
Thus, when $\rho$ is close to one, persistent writes can accumulate to a large magnitude despite asymptotic stability.

Mamba-3 introduces an additional phase degree of freedom. Consider the
corresponding rotating recurrence
\begin{equation}
    z_t
    =
    \rho e^{i\theta} z_{t-1} + u,
    \qquad \rho<1.
\end{equation}
For $z_0=0$,
\begin{equation}
    z_t
    =
    \frac{1-(\rho e^{i\theta})^t}
         {1-\rho e^{i\theta}}u,
\end{equation}
so that
\begin{equation}
    \boxed{
    \|z_\infty\|
    =
    \frac{\|u\|}
         {|1-\rho e^{i\theta}|}
    }.
    \label{eq:rotation_accumulation_bound}
\end{equation}
Without rotation, $\theta=0$, this reduces to
$\|u\|/(1-\rho)$, which can become large when $\rho\approx1$.
With nonzero phase, however, the accumulation is controlled by the complex gap $|1-\rho e^{i\theta}|$ rather than the radial gap $1-\rho$ alone. Consequently, a near-unit retention modulus does not necessarily imply large state growth: rotation can modulate constructive accumulation without changing the radial retention factor.

Geometrically, purely radial dynamics repeatedly accumulate collinear
contributions,
\[
u,\ \rho u,\ \rho^2u,\ldots,
\]
whereas rotating dynamics accumulate
\[
u,\ \rho e^{i\theta}u,\ \rho^2e^{i2\theta}u,\ldots,
\]
whose directions vary over time.

\subsection{Householder Reflections for Rotation}
\label{app:math:ckda_derivations}

For a unit-norm key $k$, setting $\beta=2$ gives the
Householder reflection $H_k=I-2kk^\top$. In the two-dimensional CKDA construction, let
\[
k=(\cos(\theta/2),\sin(\theta/2))^\top,
\qquad
D_\alpha=\operatorname{Diag}(\alpha,1),
\qquad -1<\alpha<0.
\]
For the chosen key $k$, the reflection matrix is
\begin{equation}
    \begin{aligned}
        H_k
        &=I-2kk^\top\\
        &=
        \begin{bmatrix}
            1-2\cos^2(\theta/2)
            &-2\cos(\theta/2)\sin(\theta/2)\\
            -2\cos(\theta/2)\sin(\theta/2)
            &1-2\sin^2(\theta/2)
        \end{bmatrix}\\
        &=
        \begin{bmatrix}
            -\cos\theta&-\sin\theta\\
            -\sin\theta&\cos\theta
        \end{bmatrix}.
    \end{aligned}
\end{equation}
Since $\alpha<0$, the diagonal gate decomposes as
\begin{equation}
    D_\alpha
    =H_{e_1}\operatorname{Diag}(-\alpha,1),
    \qquad
    H_{e_1}=
    \begin{bmatrix}
        -1&0\\
        0&1
    \end{bmatrix}.
\end{equation}
The two reflections compose to a rotation:
\begin{equation}
    H_kH_{e_1}
    =
    \begin{bmatrix}
        -\cos\theta&-\sin\theta\\
        -\sin\theta&\cos\theta
    \end{bmatrix}
    \begin{bmatrix}
        -1&0\\
        0&1
    \end{bmatrix}
    =R(\theta).
\end{equation}
Therefore,
\begin{equation}
    T_{\alpha,\theta}
    =H_kD_\alpha
    =R(\theta)\operatorname{Diag}(-\alpha,1).
\end{equation}

\section{Implementation Details}
\subsection{State Tracking Tasks}
\label{app:toys}


Following the setup of \citet{lahoti2026mamba}, we use a sequence-length curriculum in which the minimum sequence length is fixed at 3, while the maximum sequence length increases linearly from 40 to 160. For Parity and Mod-3, this increase spans the first 6,000 optimization steps, after which the maximum remains fixed at 160; for modular arithmetic, it spans all 10,000 steps. For each task, training lasts for $10^4$ optimization steps with a batch size of 256. We use one-layer models for Parity and Mod-3, and three-layer models for the modular-arithmetic tasks. The state dimension is fixed at 64. We optimize all models with AdamW, fix $d_{\mathrm{model}}=64$, and sweep eight learning rates logarithmically spaced between $10^{-4}$ and $10^{-2}$. We select the hyperparameter configuration with the highest validation accuracy and report its performance at sequence length 224. 

To investigate the effect of damping regime on state-tracking performance, we construct three variants based on our second-order reparameterization in Eq.~\ref{eq:general_second_order}. Specifically, we set $\kappa_t=-\nu_t^2$, $\kappa_t=0$,
and $\kappa_t=r_t^2$ for the underdamped, critically damped,
and overdamped variants, respectively. For the overdamped
variant, we parameterize $r_t=c a_t\tanh(\nu_t)$ with
$0<c\leq1$, ensuring $0\leq\kappa_t<a_t^2$. These constraints enforce the corresponding spectral families throughout training. The remaining architecture and task-specific training protocol are identical across the three variants. 

\subsection{Oscillation across Language Tasks}
\label{app:imp:across}
We evaluate official pretrained Mamba-3 1.5B SISO and MIMO models\footnote{\url{https://huggingface.co/state-spaces}} on factual relations from the Relations dataset~\citep{hernandez2024linearity} and on LAMBADA~\citep{paperno2016lambada}. Following the use of layer windows for interventions in Mamba~\citep{sharma2024locating}, we apply phase removal within contiguous layer windows, sweeping both the window size and its position across model depth. Within each selected window, we set the rotation angles to zero. We evaluate the same examples with and without the intervention, reporting the accuracy difference in percentage points.

\subsection{From-Scratch Pretraining}
\label{app:pretrain}
We train Mamba-3 MIMO 187M and 444M models from scratch on 100B tokens of the FineWeb-Edu dataset~\citep{penedo2024fineweb} with the Llama-3.1 tokenizer~\citep{grattafiori2024llama} at a 2K context length. The complete training configuration is summarized in Table~\ref{tab:pretraining_details}. 

To assess length extrapolation, we evaluate our pretrained models on LongBench-E~\citep{bai2024longbench} and PG-19~\citep{rae2019compressive}. For PG-19, we follow prior window-based evaluations~\citep{ben2025decimamba,ye2025longmamba}.
For each context length \(C\in\{1,2,4,8,16,32,64\}\times 10^3\) tokens, we include all
books containing \(L>C\) tokens. From each book, we extract length-\(C\) windows beginning at positions \(k\Delta\), where \(k=0,1,\ldots\), \(k\Delta<L-C\), and
\(\Delta=\max(10,\lfloor(L-C)/10\rfloor)\). We reset the recurrent state at the start of each window and score only its final 100 tokens, conditioned on the preceding tokens in that window. We report perplexity as the exponential of the mean negative log-likelihood over all scored tokens.

To test whether oscillatory dynamics improve performance beyond the 2K training context across datasets, we evaluate our pretrained models on CodeParrot~\citep{tunstall2022natural}, MATH-Hard~\citep{hendrycksmath2021},
TriviaQA~\citep{joshi2017triviaqa}, and SlimPajama~\citep{cerebras2023slimpajama}. Following \citet{grazzi2025unlocking}, we tokenize each dataset with the pretraining tokenizer, concatenate its documents with EOS separators, and split the resulting stream into 4096-token sequences. We reset the recurrent state at the start of each
sequence and carry it through four consecutive 1024-token position bins. For each bin, we exponentiate the mean negative log-likelihood of its tokens across sequences to obtain perplexity. The first two bins cover positions within the 2048-token training context, while the last two evaluate positions beyond it.

To test the zero-shot language modeling capabilities, we use lm-evaluation-harness~\citep{eval-harness} to evaluate our pretrained models on LAMBADA~\citep{paperno2016lambada}, HellaSwag~\citep{zellers2019hellaswag}, PIQA~\citep{bisk2020piqa}, Arc-Easy/Arc-Challenge~\citep{clark2018think}, WinoGrande~\citep{sakaguchi2021winogrande} and OpenBookQA~\citep{mihaylov2018can}.

\begin{table*}[h]
    \centering
    \caption{
        Implementation details for from-scratch pretraining on FineWeb-Edu 100B tokens.
    }
    \label{tab:pretraining_details}
    \small
    \setlength{\tabcolsep}{6pt}
    \renewcommand{\arraystretch}{1.0}

    \begin{tabular}{l|cc}
        \toprule
        \textbf{parameters}
        & \textbf{187M}
        & \textbf{444M} \\
        \midrule

        \multicolumn{3}{l}{\textit{Model Architecture}} \\
        Layers
            & 12 & 24 \\
        $d_{\mathrm{model}}$
            & 768 & 1024 \\
        $d_{\mathrm{state}}$
            & 128 & 128 \\
        Expansion Factor
            & 2 & 2 \\
        Head Dimension
            & 64 & 64 \\
        Intermediate Dimension
            & 1280 & 1792 \\
        Number of Groups 
            & 1 & 1 \\
        MIMO Rank
            & 4 & 4 \\
        Chunk Size
            & 16 & 16 \\
        Rotary Channel Fraction
            & $0/0.5$ & $0/0.5$ \\

        \midrule
        \multicolumn{3}{l}{\textit{Optimization}} \\
        Optimizer
            & AdamW & AdamW \\
        Peak LR
            & $3\times10^{-3}$ & $1.5\times10^{-3}$ \\
        Minimum LR
            & $10^{-5}$ & $10^{-5}$ \\
        LR Schedule
            & Cosine & Cosine \\
        Warmup Steps Percentage
            & 0.375\% & 0.375\% \\
        $\beta_1,\beta_2$
            & $0.9,\,0.95$ & $0.9,\,0.95$ \\
        Weight Decay
            & 0.1 & 0.1 \\
        Gradient Clip
            & 1.0 & 1.0 \\
        Global Batch Size
            & 256 & 256 \\

        \midrule
        Precision
            & \makecell{BF16 (FP32 params/\\ optimizer states)}
            & \makecell{BF16 (FP32 params/\\ optimizer states)} \\

        \bottomrule
    \end{tabular}
\end{table*}

\subsection{Pretrained Model Phase Intervention}
\label{app:eval}

We conduct the evaluations in Sec.~\ref{eval} using official pretrained Mamba-3 1.5B SISO and MIMO checkpoints. 

\noindent\textbf{Learned Phase Dynamics.}
To visualize phase dynamics during natural-language processing, we analyze the first 16,000 tokens of PG-19 validation books and record the angle projections and time steps. We reconstruct the cumulative phase
using the official phase-accumulation kernel at its native numerical precision. For each model, we examine all phase channels and select trajectories near the 10th, 50th, and 99th percentiles of phase travel among channels that complete at least one cycle. These trajectories illustrate a range of slow to fast rotary modulation on natural-language inputs.

\noindent\textbf{Interventions on Phase Structure.}
We apply four interventions using the PG-19 test protocol described above. Each intervention acts across all layers and heads throughout the sampled window. At each context length, native and intervened runs use the same books, window positions, and scored tokens.

\emph{Decay permutation} reassigns decay coefficients across heads within each layer while preserving each head's time step:
$$
\widetilde{(A\Delta)}_{\ell,h,t}
=
A_{\ell,\pi_\ell(h),t}\Delta_{\ell,h,t},
$$
where $\pi_\ell$ is a fixed cyclic permutation of the heads in layer $\ell$. Phase dynamics remain unchanged.

\emph{Phase shuffling} tests whether phase increments must remain assigned to their original rotary pairs. At each layer, head, and token, we permute all rotary pair assignments before phase accumulation:
\[
\widetilde{\delta\phi}_{\ell,h,j,t}
=
\delta\phi_{\ell,h,\sigma_{\ell,h,t}(j),t}.
\]
Thus, pair $j$ receives the increment originally computed for pair $\sigma_{\ell,h,t}(j)$. The permutation preserves the set of phase increments within each head and token but changes which pair accumulates each increment. Time steps and radial decay remain unchanged.

\emph{Phase removal} sets the raw angle projections to zero, so all phase increments vanish and the resulting rotations are identities. 

\emph{Phase reversal} negates each phase increment before
accumulation, reversing the direction of rotation.

We report token-aggregated perplexity and the ratio $\mathrm{PPL}_{\mathrm{intervened}}/ \mathrm{PPL}_{\mathrm{native}}$ at each context length.

\noindent\textbf{Phase and Recurrent State Accumulation.}
We compare native dynamics and phase removal on PG-19
validation books with at least 64,000 tokens. For each book
and condition, we process the first 64,000 tokens from a
zero recurrent state using teacher forcing and the recurrent
decoding kernel. Measurements at 16K and 64K use the same
uninterrupted trajectory.

In the kernel's state coordinates, we write
\[
S_t=P_t+U_t,
\qquad P_t=\alpha_t S_{t-1},
\]
where $U_t$ is the input contribution. We measure propagation--write
alignment as
\[
c_t=
\frac{\langle P_t,U_t\rangle_F}
{\|P_t\|_F\|U_t\|_F+10^{-12}},
\]
using all state coordinates. We average $c_t$ over
zero-indexed positions $[2000,L)$, take the median across
the $24\times64$ layer--head pairs, and average equally
across books. The RMS state norm is
\[
\mathrm{RMS}(L)
=
\frac{1}{N}\sum_{b=1}^{N}
\sqrt{
\operatorname{median}_{\ell,h}
\left[
\frac{1}{L-2000}
\sum_{t=2000}^{L-1}
\left\|S^{(b)}_{\ell,h,t}\right\|_F^2
\right]
}.
\]
Here, $N$ is the number of books, $L$ is the evaluated
context length, and $S^{(b)}_{\ell,h,t}$ denotes the
recurrent state of layer $\ell$ and head $h$ at token
position $t$ in book $b$.
The Frobenius norm includes all state coordinates
without dimension normalization. Both statistics exclude
the first 2,000 tokens to reduce initialization transients.

\noindent\textbf{Phase Consistency Between Writing and Readout.}
We selectively remove the rotary transformation from the
write vector $B$, the readout vector $C$, or both, only
during the final 100 prediction steps of each 64,000-token
window. The preceding context is processed with native
phase dynamics, and the recurrent state is carried into
the intervention region without resetting.
Removing phase replaces the corresponding rotation with
the identity while retaining the write and readout vectors.
The intervention applies to all layers, heads, and rotary
pairs, including all ranks in MIMO.
We evaluate all four conditions on all PG-19 test books, scoring only the final 100 target tokens.


\subsection{LongBench-E Task Abbreviations}

For compact presentation, we use the following abbreviations for
LongBench-E tasks throughout the paper.

\begin{table}[htbp]
    \centering
    \caption{Task-name abbreviations used for LongBench-E evaluation.}
    \label{tab:longbench_abbreviations}
    \small
    \setlength{\tabcolsep}{7pt}
    \renewcommand{\arraystretch}{1.08}

    \begin{tabular}{ll@{\qquad}ll}
        \toprule
        \textbf{Abbr.} & \textbf{Task}
        & \textbf{Abbr.} & \textbf{Task} \\
        \midrule
        2WM & 2WikiMQA
            & GR  & GovReport \\
        HQA & HotpotQA
            & LCC & Long Code Completion \\
        MQA & MultiFieldQA-en
            & MN  & MultiNews \\
        PC  & Passage Count
            & PR  & PassageRetrieval-en \\
        QA  & Qasper
            & RB  & RepoBench-P \\
        SS  & SAMSum
            & TR  & TREC \\
        TQA & TriviaQA
            &     & \\
        \bottomrule
    \end{tabular}
\end{table}

\FloatBarrier
\Needspace{0.55\textheight}
\section{Additional Results}

\subsection{Phase Structure Induced by Oscillation}
\label{app:solution}

\begin{figure}[H]
    \centering
    \includegraphics[width=0.9\linewidth]{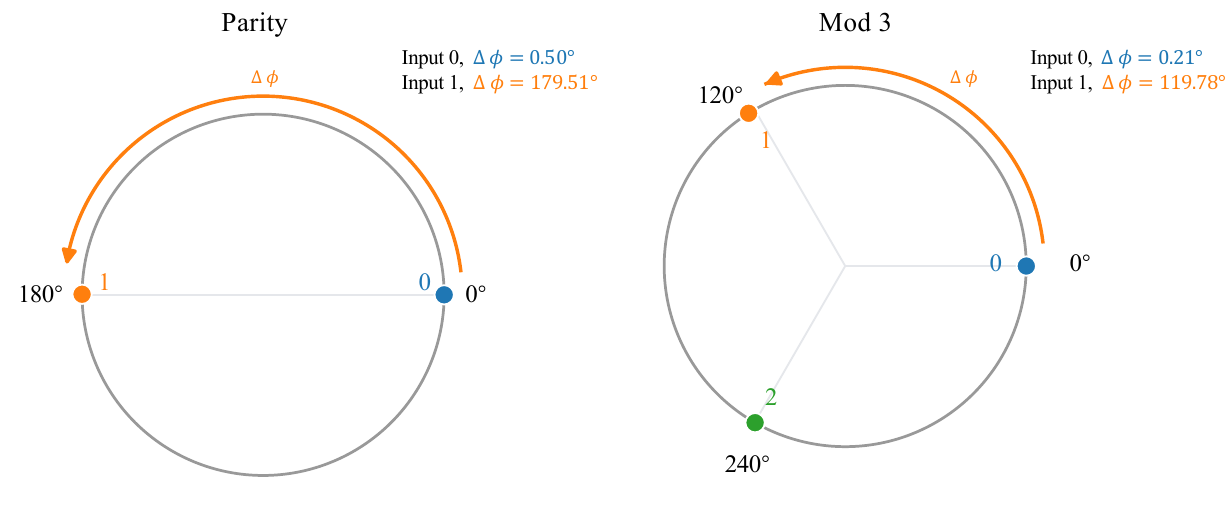}
    \caption{Task-aligned phase structure in selected two-dimensional
    state pairs. Input 0 induces little phase change, whereas input 1
    induces a rotation close to \(180^\circ\) for parity and
    \(120^\circ\) for mod-3.}
    \label{fig:task_aligned_phase_structure}
\end{figure}

Figure~\ref{fig:task_aligned_phase_structure} illustrates task-aligned phase structure in selected two-dimensional neural state pairs. For parity, one pair changes phase by \(0.50^\circ\) on input 0 and \(179.51^\circ\) on input 1. For mod-3, a selected pair shows corresponding changes of \(0.21^\circ\) and \(119.78^\circ\). These angles are close to the increments of an ideal phase code for counting ones modulo \(m\): input 0 leaves the phase unchanged, while input 1 advances it by \(360^\circ/m\). Repeated updates would then place the two parity classes near \(0^\circ\) and \(180^\circ\), or the three mod-3 classes near \(0^\circ\), \(120^\circ\), and \(240^\circ\).

This interpretation gives a concrete role to input-dependent oscillation in state tracking. In a direct rotary implementation, the phase increment must depend on the current symbol: a fixed rotation on every step would also advance the phase on input 0 and would not implement this particular count-of-ones code. The observed pairs are therefore consistent with the model using conditional rotations to represent modular state, although they do not establish that every state pair follows the same geometry.

\subsection{Length Generalization on Other Datasets}
\label{app:results:generalization}

\begin{figure}[h]
    \centering
    \includegraphics[width=\linewidth]{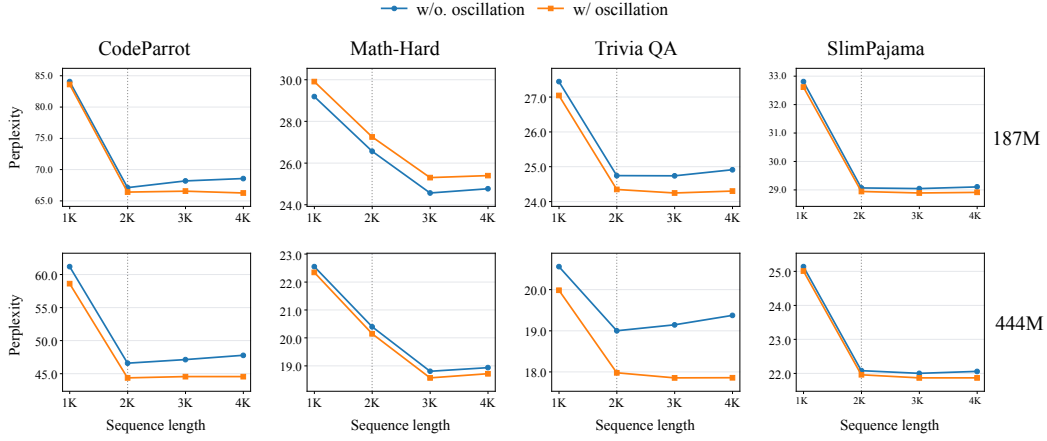}
    \caption{
        Perplexity across context lengths on CodeParrot,
        MATH-Hard, TriviaQA, and SlimPajama for models
        with and without oscillatory dynamics.
        Rows correspond to the 187M (top) and 444M (bottom)
        models. Dotted vertical lines mark the 2K training
        context length.
    }
    \label{fig:other_datasets}
\end{figure}

As shown in Fig.~\ref{fig:other_datasets}, oscillatory
models generally achieve lower perplexity, with more
consistent improvements at the larger model scale.
At 187M parameters, the oscillatory model has lower
perplexity on CodeParrot, TriviaQA, and SlimPajama,
but higher perplexity on MATH-Hard.
At 444M parameters, it has lower perplexity on all
four datasets across the evaluated context lengths.
The differences beyond the 2K training context are
most apparent on CodeParrot and TriviaQA, while those
on SlimPajama remain small. These results provide
additional evidence of improved performance at
evaluated lengths up to 4K, with benefits that vary
across datasets and model scales.

\subsection{Training Dynamics}
\label{app:training_dynamics}

\begin{figure}[h]
    \centering
    \includegraphics[width=\linewidth]
    {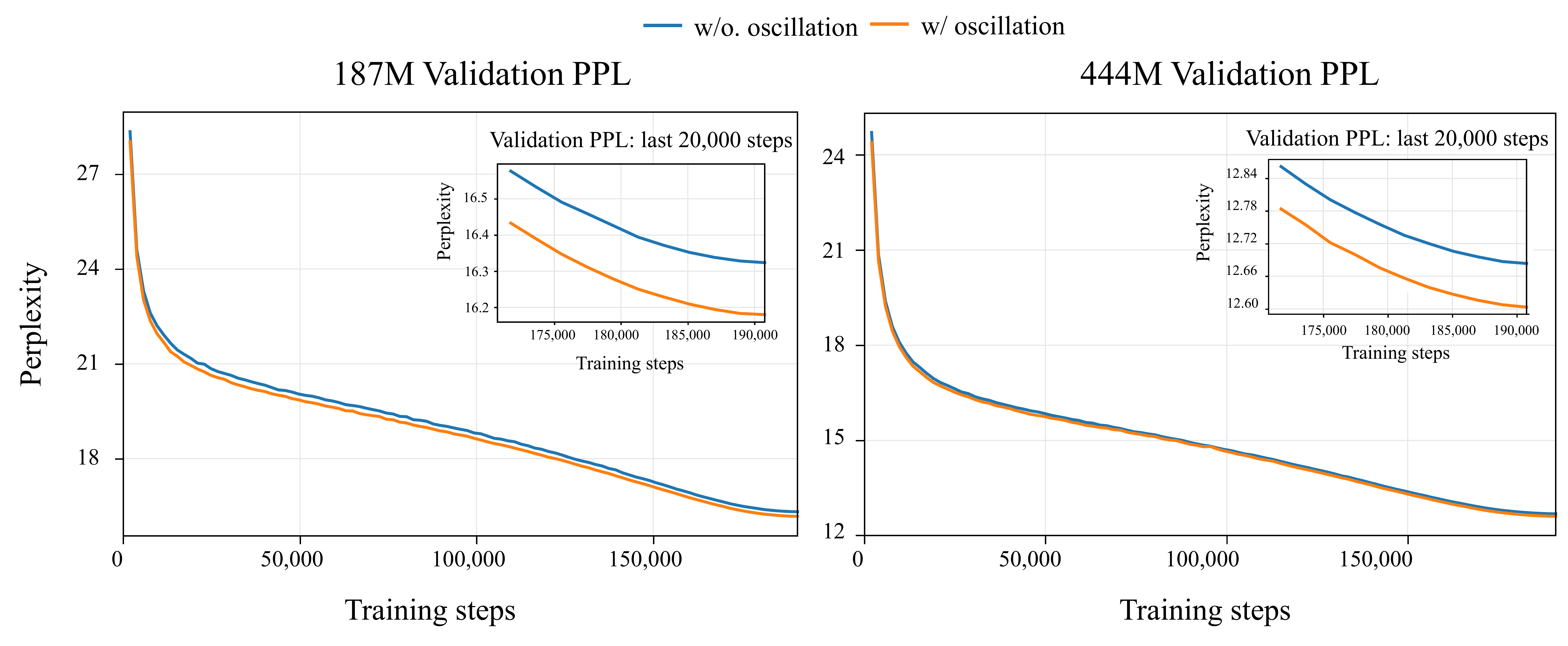}
    \caption{
        Validation perplexity throughout pretraining.
        Training trajectories of Mamba-3 187M and 444M with
        and without oscillatory dynamics.
        Insets show the final 20,000 training steps.
    }
    \label{fig:mambatrain}
\end{figure}

We further examine how the advantage of oscillatory dynamics
develops throughout pretraining.
Figure~\ref{fig:mambatrain} shows the validation perplexity
trajectories of the 187M and 444M models.
At both model scales, the oscillatory variant reaches lower
validation perplexity toward the end of training.

\begin{figure}[h]
    \centering
    \includegraphics[width=\linewidth]
    {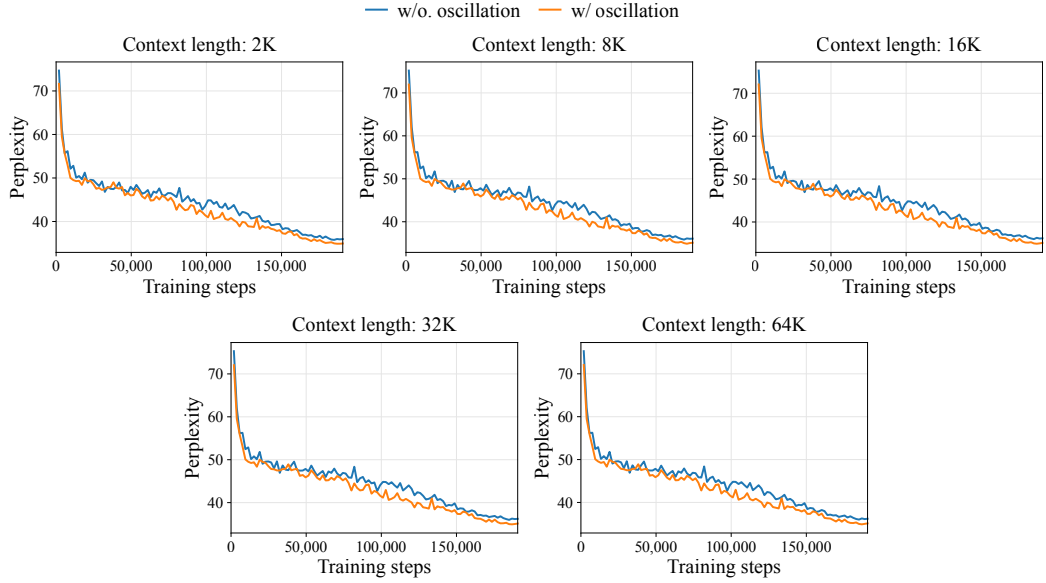}
    \caption{
        PG-19 perplexity throughout pretraining for
        Mamba-3 187M.
        PG-19 perplexity is evaluated at context lengths from
        2K to 64K at different training checkpoints.
        The oscillatory variant achieves lower perplexity across
        all evaluated context lengths toward the end of training.
    }
    \label{fig:pg19trainmamba187m}
\end{figure}

\begin{figure}[h]
    \centering
    \includegraphics[width=\linewidth]
    {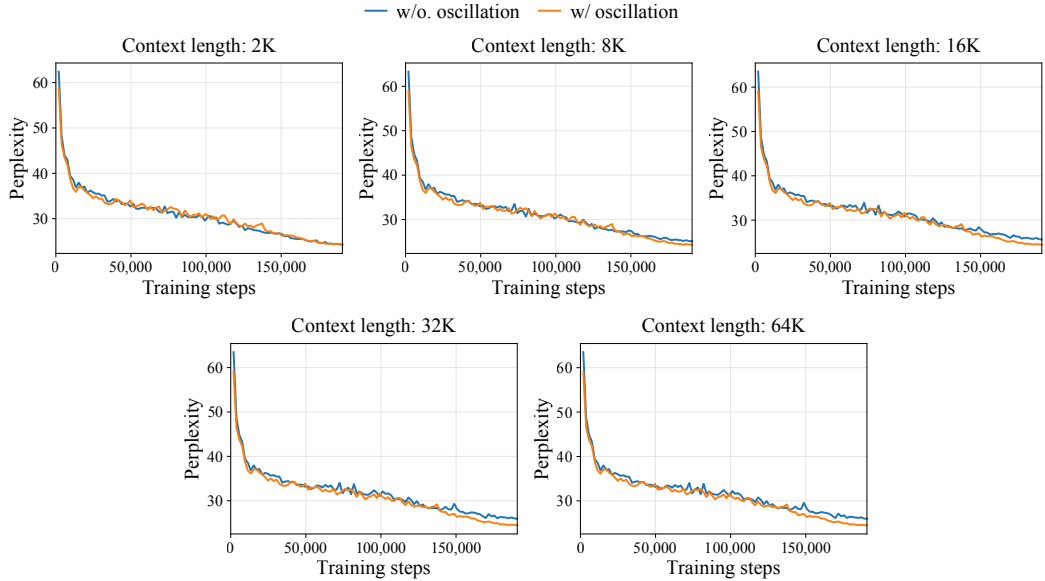}
    \caption{
        PG-19 perplexity throughout pretraining for
        Mamba-3 444M.
        PG-19 perplexity is evaluated at context lengths from
        2K to 64K at different training checkpoints.
        The oscillatory variant achieves lower perplexity across
        all evaluated context lengths toward the end of training,
        with a more pronounced separation at longer contexts.
    }
    \label{fig:pg19trainmamba444m}
\end{figure}

Figures~\ref{fig:pg19trainmamba187m}
and~\ref{fig:pg19trainmamba444m} track PG-19 perplexity at
different training checkpoints across context lengths from
2K to 64K.
At the end of training, the oscillatory variant achieves lower
perplexity at every evaluated context length for both model sizes.
The separation is more pronounced at longer context lengths,
particularly for the larger 444M model: while the two variants
remain close at short contexts, their perplexity gap becomes
substantially larger at 16K--64K.
These trajectories suggest that the benefit of oscillatory
dynamics becomes more pronounced for larger models operating
over long contexts.

\end{document}